\PassOptionsToPackage{dvipsnames}{xcolor}

\documentclass[conference]{IEEEtran}
\usepackage[square,numbers,sort&compress]{natbib}

\usepackage[font=small]{subcaption}
\usepackage[font=small]{caption}  

\usepackage{float}
\usepackage{microtype}
\usepackage{cancel}
\usepackage{multicol}
\usepackage{multirow}
\usepackage{dcolumn}
\usepackage{amsmath,amsthm,amssymb}
\usepackage{amsfonts}
\usepackage{array}
\usepackage{booktabs}
\usepackage{xcolor}
\usepackage{graphicx}
\usepackage{pifont}  
\usepackage{hyperref}
\usepackage[capitalize]{cleveref}
\usepackage{url}
\usepackage{color, colortbl}
\definecolor{Gray}{gray}{0.95}  

\newcommand{\bx}{\mathbf{x}}
\newcommand{\by}{\mathbf{y}}
\newcommand{\bX}{\mathbf{X}}
\newcommand{\bY}{\mathbf{Y}}
\newcommand{\ba}[0]{\mathbf{a}}

\newcommand{\bo}{\mathbf{o}}

\newcommand{\bi}{\mathbf{i}}
\newcommand{\bz}{\mathbf{z}}
\newcommand{\bZ}{\mathbf{Z}}

\newcommand{\E}{\mathbb{E}}

\newcommand{\bmu}{\boldsymbol\mu}
\newcommand{\bsigma}{\boldsymbol\sigma}

\newcolumntype{H}{>{\setbox0=\hbox\bgroup}c<{\egroup}@{}}
\newcommand{\mypara}[1]{\vspace{1mm}\noindent\textbf{#1}:}
\definecolor{gold}{rgb}{0.83, 0.69, 0.22}
\newcommand{\cmark}{\green{\ding{51}}}  
\newcommand{\xmark}{\red{\ding{55}}}  
\newcommand{\red}[1]{\textcolor{red}{#1}}
\newcommand{\yellow}[1]{\textcolor{gold}{#1}}
\newcommand{\orange}[1]{\textcolor{orange}{#1}}
\newcommand{\green}[1]{\textcolor{ForestGreen}{#1}}
\newcommand{\blue}[1]{\textcolor{blue}{#1}}

\usepackage{xspace}

\newcommand{\ours}[0]{{CfO}}
\newcommand{\oursfull}[0]{{Contingencies from Observations}}

\DeclareMathOperator{\argmax}{argmax}
\DeclareMathOperator{\argmin}{argmin}

\definecolor{mydarkblue}{rgb}{0,0.08,0.45}
\hypersetup{ %
    pdftitle={},
    pdfauthor={},
    pdfsubject={},
    pdfkeywords={},
    pdfborder=0 0 0,
    pdfpagemode=UseNone,
    colorlinks=true,
    linkcolor=mydarkblue,
    citecolor=mydarkblue,
    filecolor=mydarkblue,
    urlcolor=mydarkblue,
    pdfview=FitH}

\begin{document}

\captionsetup{justification=justified,singlelinecheck=false}

\title{\vspace{3mm}{\huge
\oursfull{}: Tractable Contingency Planning with Learned Behavior Models
\vspace{-5pt}
}} 

\author{\IEEEauthorblockN{Nicholas Rhinehart${}^*$, Jeff He${}^*$, Charles Packer, Matthew A. Wright \\ 
Rowan McAllister, Joseph E. Gonzalez, Sergey Levine}
 \IEEEauthorblockA{%
    UC Berkeley
    \vspace{-10pt}
  }}
\maketitle

\begin{abstract}
Humans have a remarkable ability to make decisions by accurately reasoning about future events, including the future behaviors and states of mind of other agents. Consider driving a car through a busy intersection: it is necessary to reason about the physics of the vehicle, the intentions of other drivers, and their beliefs about your own intentions. If you signal a turn, another driver might yield to you, or if you enter the passing lane, another driver might decelerate to give you room to merge in front. Competent drivers must plan how they can safely react to a variety of potential future behaviors of other agents before they make their next move. This requires contingency planning: explicitly planning a set of conditional actions that depend on the stochastic outcome of future events. Contingency planning outputs a policy that is a function of future timesteps and observations, whereas standard model predictive control-based planning outputs a sequence of future actions, which is equivalent to a policy that is only a function of future timesteps. In this work, we develop a general-purpose contingency planner that is learned end-to-end using high-dimensional scene observations and low-dimensional behavioral observations. We use a conditional autoregressive flow model to create a compact contingency planning space, and show how this model can tractably learn contingencies from behavioral observations. We developed a closed-loop control benchmark of realistic multi-agent scenarios in a driving simulator (CARLA), on which we compare our method to various noncontingent methods that reason about multi-agent future behavior, including several state-of-the-art deep learning-based planning approaches. We illustrate that these noncontingent planning methods fundamentally fail on this benchmark, and find that our deep contingency planning method achieves significantly superior performance. 
\end{abstract}

\section{Introduction}

The ability of humans to 
anticipate, probe, and plan to react to what other actors could do or want
is central to many social tasks that humans find simple but AI systems find difficult \citep[]{russell2019human}. Effectively driving a car,
for example, generally requires (1) predictive dynamics models of how cars move, (2) predictive models of driver reactions,
and (3) the ability to plan around and resolve uncertainty about other drivers' intentions.
Autonomous systems that operate in multi-agent environments typically satisfy 1--3 with separate modules that decouple prediction and planning, yet in practice the modules are highly dependent on accurate perception, an unsolved task \citep{thrun2006stanley,urmson2008autonomous,paden2016survey}.
Learning-based end-to-end systems that navigate complex multi-agent environments from raw sensory input have been the focus of recent work \citep{rhinehart2020deep,zeng2019end,filos2020can}, but do not satisfy (3), because they do not model the behaviors of the other agents.

A potential approach to building a system is to first build models to forecast other agents' behavior, and then construct ``open-loop'' action plans using these models \cite{thompson2009probabilistic,ziebart2009planning}.
However, this fails to account for the co-dependency between robot actions and environment.
Specifically, it ignores how other agents would react to robot actions and how the robot's future actions should be different depending on what the other agents do. This can lead to underconfident behavior, known as the ``frozen robot problem'' \citep{trautman_unfreezing_2010}. Thus, in many real-world situations, decoupling the tasks of forecasting human behavior and planning robot behavior is a poor modeling assumption. A resolution to this problem can be achieved through \emph{contingency planning} \citep{hardy2013contingency,zhan2016non,galceran2017multipolicy,fisac2019hierarchical},  which produces a plan that is adaptive to future behaviors of other agents or the environment. This is equivalent to planning a closed-loop policy.

\begin{figure}[t]
    \centering
    \begin{subfigure}[b]{\linewidth}
    \includegraphics[width=\linewidth]{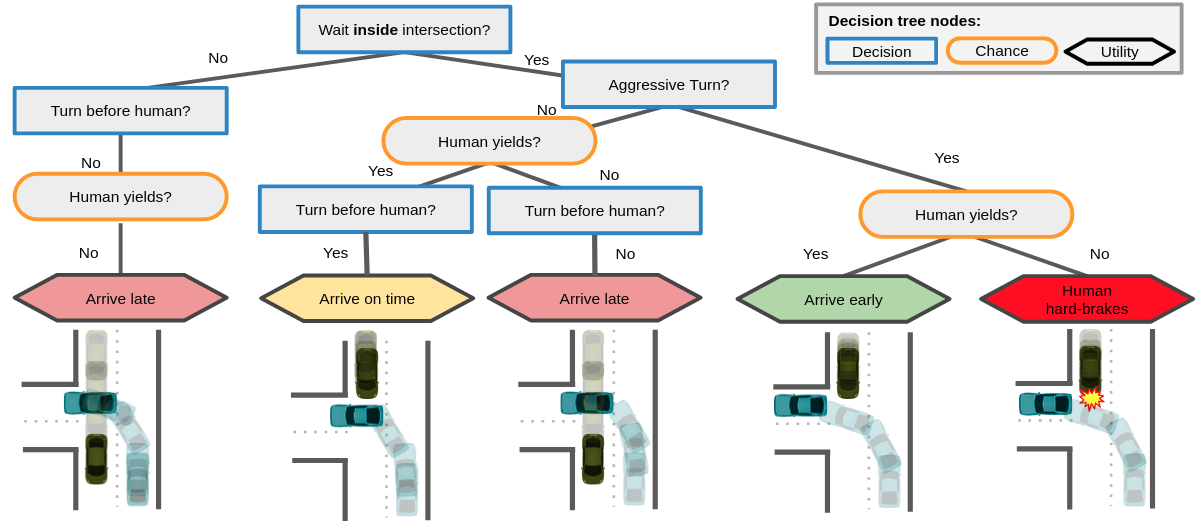}
    \end{subfigure} \\
        \caption{\small A robot's unprotected left, in which it does not have the right of way in the intersection and must either wait for the oncoming car to either pass or yield before turning. An unprotected left is a realistic multi-agent scenario that requires contingent behavior in order to achieve optimal outcomes, as the best \emph{a priori} robot trajectory suboptimally waits outside the intersection (left branch), i.e. the best sequence of noncontingent future decisions is (Wait Outside, Turn After Human), because the robot cannot be guaranteed to be able to traverse a fixed spatiotemporal path due to uncertainty about the human's intention. However, a contingent planner can achieve a better result: the optimal plan is to enter the intersection in order to observe if the other driver will yield to the robot, and either turn or wait contingent upon the other driver's decision. \textbf{Top}: The decision tree of robot and human behavior. Our approach, \oursfull, uses behavioral observations sampled from decision trees as training data to learn a behavioral model used in a contingent planning objective. \textbf{Bottom:} Conceptual diagrams of the outcomes at each leaf. Videos, code, and trained models are available at our website: {\small \url{https://sites.google.com/view/contingency-planning/home}}. }\label{fig:unprotected_left_didactic} 
        \vspace{-1.5em}
\end{figure}

\newcommand{\Prediction}[0]{\textcolor{orange!100}{Prediction}\xspace}
\newcommand{\Planning}[0]{\textcolor{blue!60}{Planning}\xspace}
\newlength{\graphsubfiglen}
\setlength{\graphsubfiglen}{.24\textwidth}
\newlength{\graphpiclen}
\setlength{\graphpiclen}{.6\graphsubfiglen}
\newcommand{\graphspace}{\hspace{0pt}}

\begin{figure*}[ht]
    \centering
    \begin{subfigure}[t]{\graphsubfiglen}
    \centering
        \includegraphics[width=\graphpiclen]{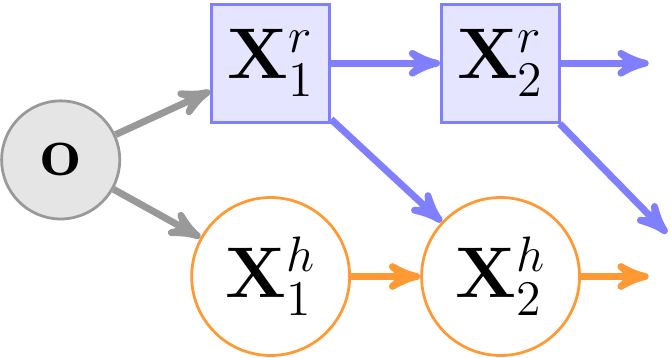}
        \caption{\textbf{Robot leader planning}. The planner does not model the human's influence on the robot, resulting in a deterministic noncontingent plan. Also known as MPC, this plans {action trajectories} independent of future behaviors of the other agents \citep{sadigh2016planning,schmerling2018multimodal,tang2019mfp}}
        \label{fig:level2}
    \end{subfigure}
    \hfill
    \begin{subfigure}[t]{\graphsubfiglen}
    \centering
        \includegraphics[width=\graphpiclen]{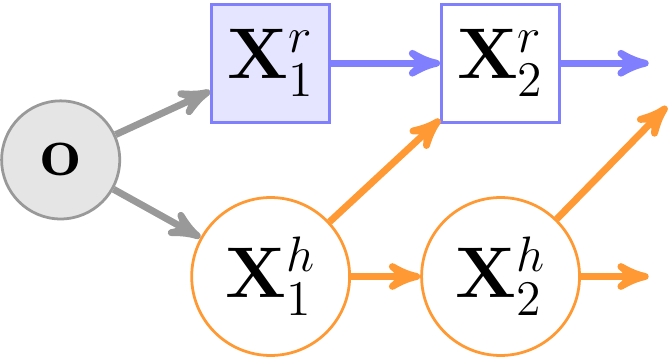}
        \caption{\textbf{Human leader planning}. The planner does not model the robot's influence on the human, which means that the robot does not model how its future actions can affect decisions of the other agents \citep{hardy2013contingency,zhan2016non}. This can result in ``underconfident'' plans.}
        \label{fig:level1}
    \end{subfigure}
    \hfill
    \begin{subfigure}[t]{\graphsubfiglen}
        \centering
        \includegraphics[width=\graphpiclen]{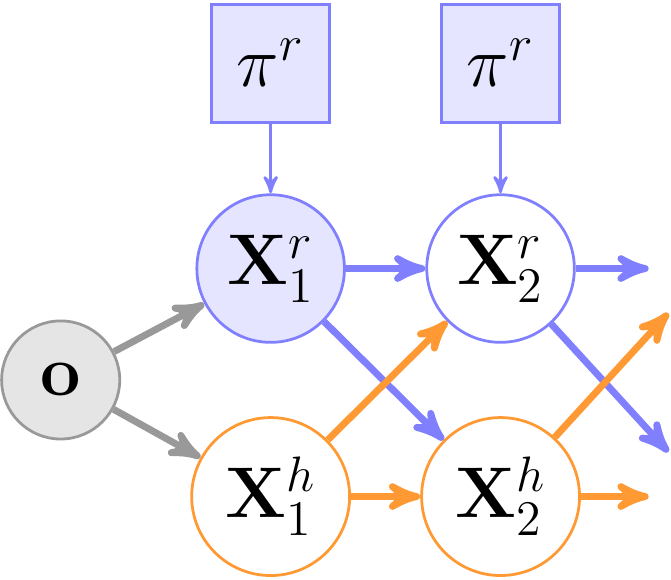}
        \caption{\textbf{Co-leader planning}. This approach plans a policy modelled to both influence and be influenced by the stochastic behavior of the human. In contrast to our method, prior contingency planning methods do not use a learned behavioral model \citep{galceran2017multipolicy,fisac2019hierarchical}.}
        \label{fig:zplanning}
    \end{subfigure}
    \hfill
         \begin{subfigure}[t]{\graphsubfiglen}
         \centering
        \includegraphics[width=\graphpiclen]{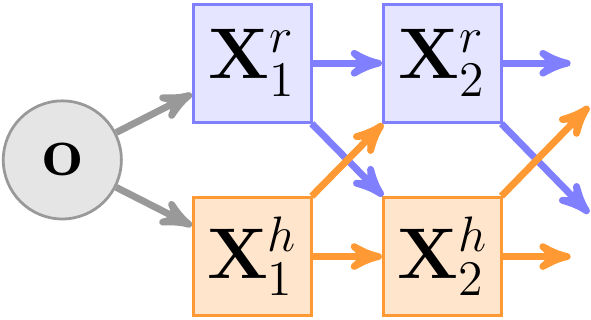}
        \caption{\textbf{Joint planning}. The planner erroneously assumes that it can control all agents, potentially resulting in ``overconfident'' behavior than can lead to unrecoverable errors. We demonstrate this phenomenon in our experiments.}
        \label{fig:overconfidentplanning}
        \vspace{-10pt}
    \end{subfigure}
    \caption{\small Various models of multi-agent interaction, based on different assumptions, as probabilistic graphical models. Circles denote random variables we can \textit{predict}, square nodes denote robot decisions we can \textit{plan}, shading indicates known values, and thick arrows represent ``carry-forward dependencies'' for visual simplicity (any two nodes connected by a chain of thick nodes has an implicit directed edge).
    }
    \vspace{-1.5em}
    \label{fig:models} 
\end{figure*}

In many partially-observed settings, optimal policies must gather information about the state of the environment (e.g. the intentions of other agents) and then act accordingly.
Consider a common scenario encountered when driving on public roads: an unprotected left-hand turn, in which the robot's goal is to turn left in the presence of another vehicle that may or may not yield, as seen in ~\cref{fig:unprotected_left_didactic}. The other driver's intention to yield must be revealed before we (the robot) commit to turning left; we must \emph{actively seek this information to resolve the uncertainty in the other driver’s intention}. Using a turn-signal indicator can provoke the other agent to reveal their intention, as can edging the robot into the intersection. Not only must we plan to be contingent upon the human, we must model the human as contingent upon us. 
Let us term this ``active contingency'' to differentiate it from ``passively contingent'' models that neglect the robot's influence on the other agents. 
In this case, we must plan to enter the intersection to indicate our desire to turn left ahead of the other driver and have a contingency planned in case the driver does not yield. 
Because the other driver may not stop, we cannot produce a single state-space plan that is guaranteed to quickly cross the intersection without potentially causing a crash, unless we wait outside the intersection. 
 
We demonstrate that autoregressive normalizing flows can explicitly represent a compact and rich space of contingent plans -- forking paths of future decisions necessary to operate successfully in environments with stochastic outcomes. We use this model class to design a deep contingency planner that scales to complex tasks with high-dimensional observations such as autonomous driving. We designed a contingency planning benchmark of common driving tasks in CARLA on which we compare our method to ablations and other methods. We show that a single, general deep contingency planning model is quite effective across several realistic scenarios, and that noncontingent planning methods fundamentally fail at these common tasks.

\vspace{-.35em}
\section{Related Work} \label{sec:related-work}
\vspace{-.25em}

Planning algorithms are central to robot navigation \citep{lavalle2006planning}, and planning under uncertainty is especially critical in uncontrolled environments like public roads.
Various choices exist for designing autonomous vehicle planning algorithms that consider other vehicles (see \citet{schwarting2018planning} for a thorough survey). We highlight some of the distinguishing concepts and design choices below, and summarize the related work in Table~\ref{table:methods}. 
In the following, we refer to \cref{fig:models}, which depicts various ways to model inter-agent dependencies.

\begin{table}[t!]
\centering
\resizebox{\linewidth}{!}{
\small
 \begin{tabular}{l c H H H c c c c c}
 \toprule
 \multirow{2}{*}{Planning Method} & \multirow{2}{*}{Contextual} & Stochastic & Continuous & High-Dimensional  & Learned  &
 \multirow{2}{*}{Contingency} \\
 & & &  State-Actions & Context & Behavior Model & \\
 \midrule
 \rowcolor{Gray}
 \citet{hardy2013contingency} & \cmark  & \cmark & \cmark & \xmark & \xmark  & \orange{Passive} \\ 
 \citet{bandyopadhyay2013intention} & \xmark & \cmark & \xmark & \xmark & \xmark  & \green{Active}  \\ 
  \rowcolor{Gray}
 \citet{xu2014motion} & \xmark & \cmark & \xmark & \xmark & \xmark  & \red{None} \\ 
 \citet{zhan2016non} & \xmark & \cmark & \cmark & \xmark & \xmark &  \orange{Passive} \\ 
  \rowcolor{Gray}
 \citet{sadigh2016planning} & \xmark & \xmark  & \cmark & \xmark & \xmark &  \red{None}    \\ 
 \citet{galceran2017multipolicy} & \cmark & \cmark & \cmark & \xmark & \xmark & \green{Active} \\
  \rowcolor{Gray}
 \citet{schmerling2018multimodal} & \xmark & \cmark  & \cmark & \xmark & \cmark &  \red{None}    \\ 
 \citet{rhinehart2020deep} & \cmark & \cmark  & \cmark & \cmark & \cmark  & \red{None} \\ 
  \rowcolor{Gray}
 \citet{zhou2018joint} & \xmark & a & \cmark & \xmark & \xmark & \red{None} \\
 \citet{fisac2019hierarchical} & \xmark & \cmark  & \xmark & \xmark & \xmark &  \green{Active} \\ 
  \rowcolor{Gray}
 \citet{zeng2019end} & \cmark & \cmark  & \cmark & \cmark & \cmark &  \red{None}  \\ 
 \citet{tang2019mfp} & \cmark & \cmark & \cmark  & \cmark & \cmark &  \red{None} \\ 
 \rowcolor{Gray} 
 \citet{cui2021lookout} & \cmark & \cmark & \cmark & \cmark & \cmark & \orange{Passive} \\ 
\citet{bajcsy2021analyzing} & \xmark & ? & \cmark & \xmark & \cmark & \orange{Passive} \\
 \midrule
 \oursfull & \cmark & \cmark  & \cmark & \cmark & \cmark   & \green{Active} \\ 
 \bottomrule
\end{tabular}}
\caption{\small Comparative summary of recent explicit planning methods for autonomous navigation tasks (ordered by publication date).
}
\label{table:methods}
\vspace{-2.0em}
\end{table}

\mypara{Learned behavior model, noncontingent planning}

Recently, fully learned planning approaches have shown promising results on autonomous navigation benchmarks and settings \cite{rhinehart2020deep,zeng2019end,filos2020can}. These approaches resemble model-based reinforcement learning (MBRL) because they model some aspect of environment dynamics in order to generate plans under reward functions. However, these methods will fail on tasks that require explicit contingency planning because they do not represent the future behavior of other agents, and therefore cannot be explicitly contingent; we demonstrate \cite{rhinehart2020deep} failing in our experiments. In the terminology of Fig.~\ref{fig:models}, ``robot leader'' methods, commonly referred to as MPC-shooting based methods, plan {action trajectories}, which means that the actions will be fixed for all possible future behaviors of the other agents \citep{sadigh2016planning,schmerling2018multimodal,tang2019mfp}. 

Modular approaches such as those used in the DARPA Grand Challenge \citep{thrun2006stanley,urmson2008autonomous} and modern industry systems \citep{paden2016survey} have the potential to be contingent, but are highly dependent on imperfect perception pipelines. We are not aware of a published demonstration of contingency planning in a pipeline approach on a realistic autonomous navigation task. 

\mypara{Passive contingency planning}
While prior work has used contingency planning for autonomous navigation, it has not used learned behavioral models.
In ``human leader'' approaches \citep{hardy2013contingency,zhan2016non,cui2021lookout}, the behavior prediction of the other agents is independent of the future behavior of the robot, which means that the robot does not model how its future actions can affect decisions of the other agents, shown \cref{fig:level1}. Unlike the active ``co-leader'' (\cref{fig:zplanning}) approach, ``human leader'' methods cannot plan to provoke other agents to reveal their intentions. 
\citet{galceran2017multipolicy} and \citet{fisac2019hierarchical} perform co-leader contingency planning with hand-crafted models (dynamics models, behavior models, reward functions) rather than learned behavior models.
Both \citet{cui2021lookout} and \citet{bajcsy2021analyzing} (a reachability analysis-based approach) only consider single-forking contingencies, whereas our method considers contingencies at every timestep, and therefore does not require determining a `branching time'.

\mypara{Model-free methods}
An alternative to using the model-based planning approaches described above is to learn a model-free {policy}, which has the capacity to implicitly represent contingent plans. Model-free imitation learning (IL) and reinforcement learning (RL) have been used to construct policies in autonomous driving environments with multi-agent interaction  \citep{dosovitskiy_carla_2017,chen2019deep,codevilla2019exploring,tang2019selfplay,palanisamy2019macad,hawke2020urban}. Model-free approaches can be successful when the data and task distribution shifts between train and test are small -- a well-trained policy will naturally mimic high-quality demonstration data exhibited by a contingent expert or evoked by a well-crafted reward function. However, if there is mismatch between the training and test rewards, the policy will produce suboptimal behavior. While goal-conditioning methods can address the suboptimality of adapting model-free methods to test-time data by serving as the representation for the test-time reward function, the goal space of the agents must be specified \emph{a priori}, requires goal labels, and precludes adapting the learned system to new types of test-time goal objectives \citep{codevilla2019exploring,hawke2020urban}.
Furthermore, in contrast to model-free methods, our method explicitly represents the planned future behavior, which offers interpretability and model introspection benefits. 

Our proposed method is both contingent and uses a fully-learned behavior model, unlike prior work that either had learned behavior models or was actively contingent, but never both, as shown in Table~\ref{table:methods}.

\vspace{-.35em}
\section{Deep Contingency Planning}
\vspace{-.25em}

\begin{figure}
\centering
\begin{subfigure}[b]{\columnwidth}
\includegraphics[width=\columnwidth]{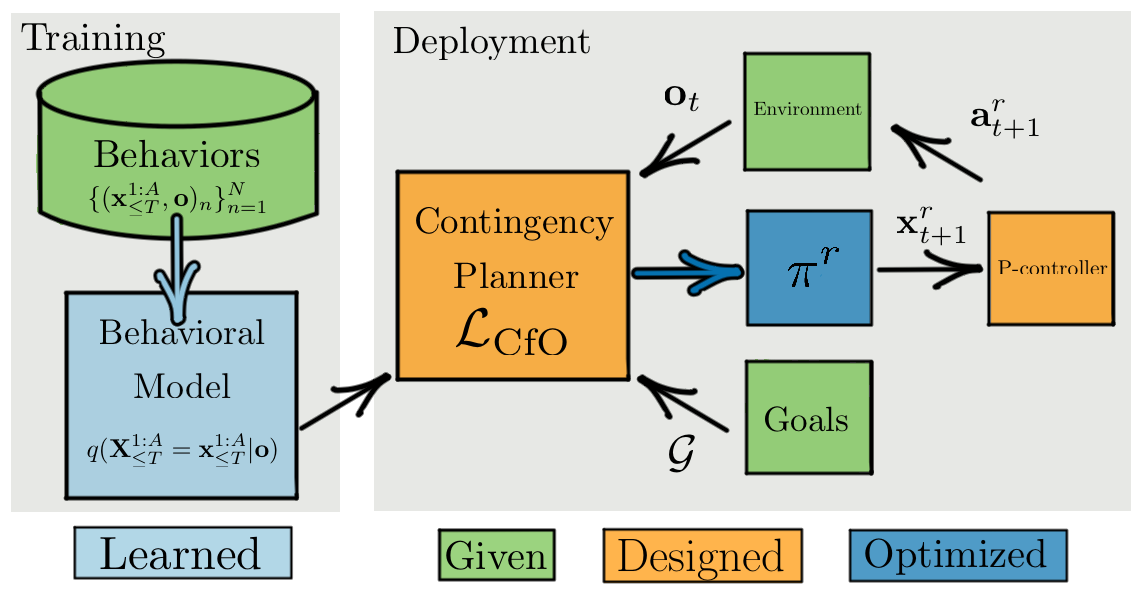}
\end{subfigure}
\caption{\small Flowchart that describes the components of our approach. A behavioral model is trained to forecast multi-agent trajectories conditioned on high-dimensional scene observations, and is used during deployment to construct a contingency planner. The contingency planner receives goals $\mathcal{G}$ from an outer navigator, as well as high-dimensional observations $\bo_t$ at timestep $t$ from the environment. The contingency planner plans a policy, which predicts the immediate next target position $\bx_{t+1}^r$. This next position is passed to a P-controller to produce the necessary action $\ba_{t+1}^r$.} \label{fig:method_overview}

\vspace{-2em}
\end{figure}

Below, we outline the deployment phase, planner design, and training phase. Our method is depicted in Fig.~\ref{fig:method_overview}. 

\mypara{Deployment} The following loop is executed during deployment: (1) the contingency planner is given a learned behavioral model, goals, and high-dimensional sensory observations as input, and outputs a multi-step plan in the form of a robot policy. (2) The planned robot policy is executed for one step to output a target position, which is (3) tracked by a proportional controller that outputs a control that includes robot steering, throttle, and braking controls. (4) The control is executed on the robot in the environment, which produces a new observation.

\mypara{Planner design} The future behavior of the robot is planned in terms of a parametric policy; a \emph{subset} of the policy's parameters are used to optimize a planning criterion.
Crucially, this policy is closed-loop -- it reacts differently to (i.e., is \emph{contingent} on) different possibilities of the future behavior of all agents (Fig.~\ref{fig:zplanning}). When the policy is open-loop, it is noncontingent, and represents the future positional trajectory of the robot, which can result in underconfident behavior (Fig.~\ref{fig:level2}). The planning criterion incorporates several components: (1) the learned behavioral model
to forecast a probability distribution of the likely behaviors of all agents in terms of positional trajectories; the PDF of this distribution is used in the planning criterion. The planning criterion also incorporates (2) goals specified as positions, and optionally incorporates (3) hard or soft constraints on joint behavior. The planner optimizes for a robot policy that has a high expected probability of joint behavior, and a high expected probability of satisfying the provided goal location and optional goal constraints; the expectation is under the other agents' sources of uncertainty.

\mypara{Model design and training} The behavioral model is trained with a dataset of trajectories of multi-agent positions paired with high-dimensional sensory observations of one of the agents. Training maximizes the likelihood of the trajectories, similar to multi-step Behavioral Cloning \cite{pomerleau1989alvinn}. Following prior IL work, we sometimes call these positions \emph{behavioral observations}, in contrast to \emph{demonstrations} (observations of state-action pairs). We assume that the behaviors in this data are goal-directed, but we do not assume that every behavior would score well under the planning criterion: some of these behaviors require navigation to different goals than those seen in training, and some of these trajectories exhibit behavior
that may be undesirable. This somewhat general assumption on the quality of the data behavior demands a planning criterion that involves terms of multi-agent trajectories, which necessitates modeling the likely trajectories of the other agents. 

\subsection{Preliminaries}

We consider multi-agent systems composed of $A\geq2$ agents (e.g., vehicles). We assume agent positions are fully observable by all agents, but the intentions of each agent (e.g.,\ intended destinations) are hidden from each other. We consider continuous positions and discrete time, and define the position of the $a$th vehicle at time $t$ by its location in a $d$-dimensional Cartesian space as $\bx^a_t \in\mathbb{R}^{d}$.
The joint positions of all vehicles are denoted  by $\bx_t\in\mathbb{R}^{A\times d}$, where a lack of superscript indicates all vehicles.
Let $\bx_t^r=\bx_t^1\in\mathbb{R}^{1\times d}$ index the position of the controlled robot vehicle, and $\bx_t^h=\bx_t^{2:A}\in\mathbb{R}^{(A-1)\times d}$ index the positions of other human-driven vehicles that the robot can influence but has no direct control over. Let $t=0$ and $\bx_0$ denote the current timestep and joint position of the multi-agent system, and let $\bx_{\leq t}$ denote future joint positions $[1, \dots, t]$.
Uppercase denotes random variables (e.g. the stochastic sequence of all multi-agent positions $\bX_{\leq t}^{1:A}$), and lowercase denotes realized values.
The current observation of the robot is denoted ${\bo} \doteq \{\bx_0, \bi_0\}$; $\bi_0$ provides high-dimensional information about the environment.
In our implementation, $\bi_0\in\mathbb{R}^{H\times W}$ is a LIDAR range map image \cite{caccia2018deep}.

\subsection{Model Design and Training Details}
\label{sec:model_design}
The planner requires a learned behavioral model that stochastically forecasts multi-agent trajectories $\bX_{\leq T}^{1:A}$ given context $\bo$, and
captures co-dependencies between agents at all timesteps. Let $q_\theta(\bX|\bo)$ denote this model.
The model factorizes into a product of autoregressive distributions parameterized by neural networks that receive observations and previous states:

\begin{align}
    q_\theta(\bX_{\leq T}^{1:A}=\bx_{\leq T}^{1:A}|{\bo})=\mathop{\prod}_{t=1}^{T}\mathop{\prod}_{a=1}^A  q^a
    (
    \bX^a_t\!=\!\bx^a_t;\phi_t^a),   \label{eq:esp_model}
    \end{align}
 where the parameters of $q^a$ are computed autoregressively in order to model reactions dependent only on past data: \mbox{$\phi_t^a=f_\theta^a(\bx_{<t}^{1:A},\bo)$}.  We restrict $q^a$ to the family of distributions that can be sampled by transforming a sample $\bz_t^a\in\mathbb R^d$ from a fixed simple base distribution chosen \emph{a priori}, $\bar{q}^a(\bz_t^a)$, to $\bx_t$, using a fixed invertible transformation of observations and learned values \mbox{$\bx_t^a=m(\bz_t^a; \phi_t^a)$}. E.g., if $q^a=\mathcal N(\cdot| \{\bmu_t^a, \bsigma_t^a\})$, then \mbox{$m(\bz_t^a; \phi_t^a)=\bmu_t^a + \bsigma_t^{a}\bz_t^a$}, and \mbox{$\bar{q}^a(\bz_t^a)=\mathcal N(\bz_t^a; 0, I)$}. Because the $\bar{q}^a$ are independent, the model represents multi-agent behavior as a learned coupling of independent variables, $\bz_t^a$. Taken together, these conditions specify that the model is a (conditional) autoregressive normalizing flow \cite{dinh2016density,papamakarios2017masked}. 

Now we show how $q_\theta$ can be used to design a compact and rich space of contingent policies that is amenable to fast optimization at planning time. First, note that the process of generating agent $a$'s position at time $t$ with $q_\theta$ can be seen as generating a position from (state-space) policy: \mbox{$q^a(\bX_t^a \mid \bx_{<t}^{1:A},\bo; \theta)$}. Next, note that given $\bz_t^a$, $m$ can be written as the application of a deterministic policy parameterized by both $\bz_t^a$ and $\theta$: \mbox{$\bx_t^a=\pi^a(\bx_{<t}^{1:A},\bo; \bz_t^a, \theta)$}. On the one hand, $q_\theta$ can be seen as a collection of stochastic policies $\left\{q^a\right\}_{a=1}^{A}$. On the other hand, $q_\theta$ can be seen as a collection of deterministic policies and policy parameter priors $\{(\pi^a, \bar{q}^a)\}_{a=1}^{A}$. At planning time, we can specify a contingent robot plan by deciding values for the robot's policy parameters, $\bz_{\leq T}^r \in \mathbb R^{2 \times T}$, rather than sampling each $\bz_t^r$ from its prior $\bar{q}^r$. Doing so results in a contingency plan that is reactive to each potential joint behavior of the other agents, $\bX_{\leq t}^{2:A}$, for $t\in[1,T]$. Because the future behaviors of the other agents are unobserved at test-time, the overall system is still stochastic given the contingency plan parameterized by $\bz^r$. This method of partially parameterizing a policy is appealing because it requires no modifications to the model. Once $q_\theta$ is trained, $\theta$ is fixed, and planning is a matter of choosing the free parameters, $\bz^r$, of $\pi^r$. Other parameterizations could certainly be used, but we found this method efficient. \cite{rhinehart2020deep,singh2020parrot} employ similar parameterizations in the noncontingent setting.

 A natural question that arises from this discussion is: which contingent plans can such a scheme actually represent? While analyzing this in the continuous case is difficult, we can provide a detailed analysis of this question in the case where all variables ($\bx$ and $\bz$) are discrete. In this case, we show in the Appendix that likely contingent plans can be expressed in terms of $\bz^r$ and can represent sequences of $n$-th best responses under the model $q_\theta$ (i.e., first-best, second-best, etc.). Key to this analysis and the general richness of the planning space is the invertibility of $m$ in $\bz_t^a$, which ensures that all $\bx_t^a \in \mathbb R^d$ are realizable outputs of $\pi^a$. In the Appendix (Fig.~\ref{fig:contingent_plan_examples}), we also give example visualizations of a single $\bz^r_{\leq T}$ resulting in high-quality plans contingent on plausible futures of other agents.

We train the model, $q_\theta$, with a dataset $\{(\bx_{\leq T}^{1:A}, \bo)_n\}_{n=1}^N$, depicted in Fig~\ref{fig:method_overview}. As previously mentioned, we assume that these behaviors are goal-directed to diverse goals, which allows the model to represent many modes of plausible behavior (e.g., turning left, turning right, traveling passively, traveling aggressively). We defer implementation details of constructing and training $q_\theta$ to Section~\ref{sec:impl}. 

\vspace{-.5em}
\subsection{Planner Design Details}\label{sec:planner_design}
\vspace{-.5em}
\mypara{Primary planning objective} Having described our learned behavioral model and a method to parameterize a contingent plan as a closed-loop policy, we turn to constructing a planning objective in order to evaluate the \emph{quality} of the planned policy in terms of how plausible its behavior is and how well it satisfies goals at test time. In our application, goals $\mathcal G=(\mathbf{g},\mathbb G)$ are provided to the robot in terms of coordinate destinations $\mathbf{g} \in \mathbb R^2$, as well as a set of constraints, $\mathbb G$, on the joint trajectory specified as an indicator function $\delta_{\mathbb{G}}(\bx)$, which is $1$ where $\bx\in\mathbb G$ and $0$ otherwise. To evaluate the quality of a plan, we take inspiration from the single-agent (non-contingent) imitative model planning objective \citep{rhinehart2020deep} by formulating planning with the model as maximum \emph{a posteriori} (MAP) estimation. Instead of MAP estimation of a robot trajectory (non-contingent plan), we perform MAP estimation of the subset of policy parameters $\bz_{\leq T}^r$ with posterior $p(\bz_{\leq T}^r|\mathcal G, {\bo})$. The resulting lower-bound objective, derived on the supplementary website, is:
 \begin{align}
    &\mathcal{L}_\text{\ours}(\pi^r_{\bz_{\leq T}})
    =  \label{eq:final_planning_criterion}\\
    &\mathop{\E}_{\bz^h \sim \mathcal{N}(0, I)} \Big[
    \log \underbrace{q_\theta(\bar{\bx}_{\leq T}|{\bo})}_{(1)~\text{prior}}\!+\!\log  \underbrace{\mathcal N\big(\bar{\bx}^r_T; \mathbf{g}, I
    \big)}_{(2)~\text{destination}}\!+\!\log \underbrace{\delta_{\mathbb G}(\bar{\bx}_{\leq T})}_{(3)~\text{constraint}} 
    \Big].\notag
 \end{align}
Eq.~\ref{eq:final_planning_criterion} consists of several terms: the (1) behavioral prior encourages the planned policy to result in joint behavior that is likely under the learned model; (2) the destination likelihood encourages the policy to end near a goal location; (3) an optional goal constraints penalizes the policy if it does not satisfy some desired constraint. The expectation results from the uncertainty due to the presence of other (uncontrolled) agents. In practice, we perform stochastic gradient ascent on Eq.~\ref{eq:final_planning_criterion} w.r.t. $\bz_{\leq T}^r$ and approximate a violated constraint ($\log \delta_{\mathbb G}(\bx_{\leq T})\!=\!-\infty$ when $\bx_{\leq T}\notin\mathbb G$) with a large negative value.

\mypara{Alternative planning objectives} Consider instead directly planning $\bx_{\leq T}^r$ under uncertain human trajectories $\hat{\bx}_{\leq T}^h$: 
 {\footnotesize
 \begin{align*}
  \mathcal L^{\text{r}}(\bx_{\leq T}^r)&=\mathop{\mathbb E}_{\hat{\bx}\sim q}\log\mathcal N(\bx_T^r;\!\mathbf{g},\!I) 
    \! +\! \log \delta_{\mathbb G}([\bx^r\!,\!\hat{\bx}^h]) 
     \! +\! \log q([\bx^r\!,\!\hat{\bx}^h]|{\bo}).
 \end{align*}
}
Planning with ${\mathcal L^{\text{r}}}$ will result in a noncontingent plan that is \emph{underconfident}, because the criterion fails to account for the fact that $\bx^r_t$ affects $\bx^h_{>t}$. This is illustrated in Fig.~\ref{fig:level1}. For example, we show in our experiments how an underconfident planner leads to a \emph{frozen robot} that prefers not to enter the intersection and indicate that it intends to turn left, because the plan -- a single fixed trajectory -- is not affected by whether or not the human driver yields to it. The best trajectory under this criterion is a passive waiting trajectory that stays outside the intersection until the human has passed. Finally, consider planning \emph{both} $\bx^r$ and $\bx^h$ (Fig.~\ref{fig:overconfidentplanning}), i.e., assuming control of both the robot and the human trajectories: 
\begin{equation*}
    \mathcal L^{\text{joint}}(\bx) =\log \mathcal N(\bx_T^r; \mathbf{g}, I) + \log \delta_{\mathbb G}(\bx) + \log q_\theta(\bx|{\bo}).
\end{equation*}
This is an \emph{overconfident} planner, because it assumes that the other agents will obey the robot's plan for them. Thus, this planner is acceptable when the human behaves as planned, but risky when the human does not. In the left-turn scenario, $\mathcal L^{\text{joint}}$ causes the robot to assume that the human will always yield to it, resulting in dangerous behavior when the human sometimes does not. Instead, if the prior of joint behavior is the only component of the planning criterion (i.e. just the final term of $\mathcal L^{\text{joint}}$), then the planner simply optimizes for the mode of the prior, which will result in an undirected likely joint behavior.

\vspace{-.25em}
\section{Experiments} \label{sec:experiments}
\vspace{-.25em}
Our main experimental hypothesis is that \emph{(i)} some common driving situations that require multi-agent reasoning can be nearly solved with our deep contingency planning approach. This hypothesis is related to the following secondary hypotheses, which state that various noncontingent planning methods cannot solve these situations: \emph{(ii)} a noncontingent learning-based underconfident planner will lead to a frozen robot in some situations, \emph{(iii)} a noncontingent learning-based planner of joint trajectories will lead to an unsafe robot in some situations, \emph{(iv)} online model-free methods will incur a significant cost of failed behaviors before learning to succeed, in contrast to our method, and \emph{(v)} oblivious planning (unable to adapt to cost function) will lead to an unsafe robot in some situations. To evaluate these hypotheses, we developed a set of common driving scenarios in which contingency planning matters.
\vspace{-.5em}
\subsection{Benchmark Scenarios}
\vspace{-.25em}
\begin{figure}[htb]
    \centering
    \begin{subfigure}[t]{\linewidth}
    \includegraphics[width=\linewidth]{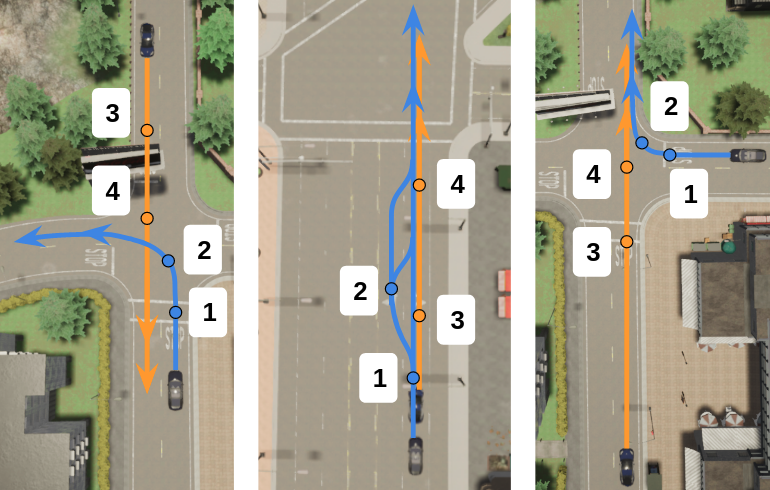}
    \end{subfigure}
        \caption{\small Overhead images for each scenario. The circles denote decision points; blue paths being robot, orange paths being ``human''. \emph{Left and Right}: The robot tries to turn. At (1), robot decides if to enter the intersection as turning signal to human. At (2), robot decides if to aggressively turn or wait.  At (3), if robot previously entered the intersection, human may or may not yield. Again at (2), if robot is waiting, it decides when to turn based on if human yielded. At (4), if robot turned aggressively and human didn’t yield, a near-collision occurs. \emph{Middle}: Robot tries to overtake. At (1), robot decides if to enter the left lane. At (2), robot decides if to overtake aggressively. At (3), human decides if to yield. Again at (2), if robot wasn't aggressive, it decides if to overtake. At (4), if robot was aggressive and human didn’t yield, a near-collision occurs.}\label{fig:task_images} 
        \vspace{-1.75em}
\end{figure}

We evaluate our method and several others on three multi-agent driving scenarios that we built in the CARLA simulator \citep{dosovitskiy_carla_2017}: a left-turn scenario similar to the one discussed in Fig.~\ref{fig:unprotected_left_didactic} and conceptually similar to that in \cite{hardy2013contingency}, a highway overtaking scenario conceptually similar to that used in prior work \citep{sadigh2016planning,fisac2019hierarchical}, and a right turn at a stop sign scenario. Images from each scenario are shown in Fig.~\ref{fig:task_images}. For each scenario, we identified 4 suitable locations, and used 1 of them as training locations and 3 as test locations. At test time, we evaluate each method 10 times in each test location, resulting in a total of 90 evaluation episodes per method ($3~\frac{\mathrm{scenarios}}{\mathrm{method}}\cdot 3\frac{\mathrm{locations}}{\mathrm{scenario}}\cdot10\frac{\mathrm{episodes}}{\mathrm{location}}$). The testing episodes use a fixed set of random seeds in the simulator. 

In the left turn scenario, the robot car needs to execute a left turn, but the oncoming vehicle may or may not yield.
In the highway overtaking scenario, the robot car must pass the car in front of it, but this requires entering the left lane. Similar to how, in the left turn scenario, the other vehicle may or may not yield, the leading car in the overtaking scenario may slow down or maintain a constant speed to facilitate the robot car's overtaking, or it may accelerate to prevent it.
In the right turn scenario, the robot car needs to execute a right turn into a lane that contains a vehicle that may or may not yield.

These scenarios, while common and easily negotiated by human drivers, have features that make them difficult for many learning-based autonomous navigation methods.
The optimal outcomes (making the turns quickly and safely, and successfully overtaking the other car) require cooperation from the human agent, but the intent of the human to cooperate is unknown.
This calls for contingent planning.
However, in order to infer the human's intent, we (the robot) must first act in a way that signals our own desire (entering the intersection or the opposite lane) and evaluate their response.

\vspace{-.5em}
\subsection{Implementation Details}\label{sec:impl}
\vspace{-.25em}
We designed these scenarios in suitable locations in the CARLA simulator \citep{dosovitskiy_carla_2017}.
We generated training data from hand-crafted policies designed to generate each of the outcomes in each scenarios:
first, the robot chooses whether to begin its maneuver (enter the intersection or the opposite lane).
If the robot begins the maneuver, the other vehicle decides whether to accommodate the it (yield to it or allow passing). Then, the robot decides whether to attempt to complete the maneuver. The robot's policy can be considered a suboptimal expert, as it sometimes generates expert paths, but other times, it generates conservative paths (by sometimes not entering the intersection or opposite lane) and risky paths (by sometimes attempting to complete the maneuver despite the other agent not yielding). Examples are given in the Appendix (Fig.~\ref{fig:oc_uc}).

We adapted the ESP model from \citep{rhinehart2019precog}, which uses $q_t^a=\mathcal N$ and we used $1.5$s of past at $10$Hz and $8$s of future at $3.75$Hz, resulting in pasts of length $15$ and futures of length $30$. In contrast to the LIDAR featurization in \citep{rhinehart2019precog}, we instead use $\bi_0\in\mathbb{R}^{H\times W}$ as a LIDAR range map image \cite{caccia2018deep}. Whereas \citep{rhinehart2019precog} performed \emph{passive forecasting} of behaviors of agents, and was not used to decide controls of a robot, our work focuses on building and executing actively contingent plans.

\noindent\textbf{Method and baselines}: We trained a \emph{single} global \ours~model, which forces the method to generalize across scenarios and locations. For comparison, we use \cite{tang2019mfp} (MFP) by following the shooting-based planning method described therein. We experimented with a global model for MFP, but were unable to achieve good performance, instead, we trained separate \emph{per-scenario MFP models}. We use the \ours~model with the contingent planning objective (Eq.~\ref{eq:final_planning_criterion}) as well as with the ``overconfident'' noncontingent planner with criterion $\mathcal{L}^\textrm{joint}$ from Section \ref{sec:model_design}. For the ``underconfident'' planner, we used the suboptimal expert to create the set of per-agent paths available at each timestep, and used $\mathcal{L}^\textrm{r}$ without the prior term, which serves as a best-case underconfident planner (in the case of a perfect model). To measure the importance of multi-agent planning vs. an oblivious planner, we apply \citep{rhinehart2020deep} by constructing a single-agent version of our model, which cannot incorporate terms that involve predicted positions of other agents. For this baseline, like MFP, we trained separate per-scenario models. Further details are given in the Appendix.

\vspace{-.5em}
\subsection{Results}
\vspace{-.25em}

\begin{table}
\centering
\resizebox{.99\columnwidth}{!}{
\begin{tabular}{l D{,}{/}{-1} D{,}{/}{-1} l D{,}{/}{-1} D{,}{/}{-1} l D{,}{/}{-1} D{,}{/}{-1}}
\toprule
& \multicolumn{2}{c}{Left Turns} &  & \multicolumn{2}{c}{Overtaking} & &  \multicolumn{2}{c}{Right Turns} \\
\cline{2-3}\cline{5-6}\cline{8-9}\vspace{-2mm}\\
Method     & \multicolumn{1}{c}{RG} & \multicolumn{1}{c}{RG*} & &  \multicolumn{1}{c}{RG}& \multicolumn{1}{c}{RG*} & &  \multicolumn{1}{c}{RG}& \multicolumn{1}{c}{RG*}\\
\midrule
Single-agent \citep{rhinehart2020deep} & {\scriptstyle 30},{\scriptstyle 30} & 16,30 & & {\scriptstyle 10},{\scriptstyle 30} & 8,30 & & {\scriptstyle 15},{\scriptstyle 30} & 2,30 \\
Noncontingent, $\mathcal{L}^\textrm{r}$ & {\scriptstyle 30},{{\scriptstyle 30}} & 0,30 & & {\scriptstyle30},{\scriptstyle30} & 0,30 & & {\scriptstyle 30},{\scriptstyle 30} & 0,30\\
Noncontingent, $\mathcal{L}^\textrm{joint}$& {\scriptstyle30},{\scriptstyle 30} & 9,30 & & {\scriptstyle29},{\scriptstyle30} & 12,30 & & {\scriptstyle23},{\scriptstyle 30} & 9,30\\
MFP \cite{tang2019mfp} & {\scriptstyle 30},{\scriptstyle 30} & 9,30 & & {\scriptstyle16},{\scriptstyle30} & 9,30 & & {\scriptstyle 9},{\scriptstyle 30} & 9,30 \\ 
\ours{} (Eq.~\ref{eq:final_planning_criterion}) (ours)& {{\scriptstyle 29}},{{\scriptstyle 30}} & \mathbf{28},\mathbf{30} & & {\scriptstyle27},{\scriptstyle30} & \mathbf{27},\mathbf{30} & & {\scriptstyle 27},{\scriptstyle30} & \mathbf{27},\mathbf{30}\\
\bottomrule
\end{tabular}
}
\caption{{\small Comparison of learning-based planning for control: Our co-Influence model (\ours{}) learns how agents behave together, and is able to control a vehicle to negotiate multi-agent driving scenarios safely and efficiently. Ablations show that alternative non-contingent planners lead to either overly conservative or unsafe trajectories.}} \label{table:evaluation} 
\vspace{-2em}
\end{table}

Prior navigational planning work uses a variety of metrics to measure the quality of a planned or predicted trajectory \citep{tang2019mfp,rhinehart2019precog}.
In this work, we evaluate our different planners in terms of closed-loop control metrics: the ability of a method to reach a goal (1) successfully, and (2) safely and efficiently.
In Table \ref{table:evaluation}, we report results in terms of the goal-reaching success rate (``RG"), which simply means that the robot vehicle eventually reached the goal region, and \emph{near-expert} success rate (``RG*''), which we define as the vehicle reaching the goal as quickly as an expert would, without any near-collisions.

Examining Table~\ref{table:evaluation}, we see evidence that supports the hypotheses mentioned at the beginning of the section.
First, \emph{(i)} the \ours{}
planner (i.e., the one that uses Eq. \ref{eq:final_planning_criterion}) has near perfect success rate in both scenarios, as well as a nearly perfect rate of near-expert success.
The high rate of completing the maneuver in a near-expert way suggests that contingency planning is helpful for efficiently navigating these complex scenarios, which is possible only by probing the human's intentions and reacting accordingly.
Next, \emph{(ii)} our ``underconfident'' noncontingent planner, while having a high success rate, has a much lower count of near-expert successes than the contingent \ours{} planner.
The reason is that the underconfidence leads to ``frozen robot'' behavior that slowed down the underconfident planner.
In the left turn scenario, for example, the underconfident planner will always wait for the oncoming vehicle to pass through the intersection before beginning its own left turn.
This means the underconfident planner missed any opportunity to negotiate the intersection as quickly as the expert did in the trials where the human vehicle would have yielded.

For \emph{(iii)}, note that the ``overconfident'' noncontingent planner always reaches the goal, but sometimes engages in unsafe behavior. Its aggressive planning means that it will try to solve the scenario quickly every time, which leads to near-collisions every time that the other vehicle does not yield.

For \emph{(iv)}, we observe the ``oblivious'' planner \cite{rhinehart2020deep} indeed performs suboptimality across scenarios, leading to more near-collisions than the other planners. Due to the suboptimality present in the training data, its planner, which reasons solely about the ego-vehicle's behavior, is insufficient in these scenarios.
Finally, for \emph{(v)}, we compare our approach to a model-free reinforcement learning (RL) baseline, Proximal Policy Optimization (PPO), on the left turn scenario.

\begin{figure}[t]
    \centering
    \includegraphics[width=.95\linewidth]{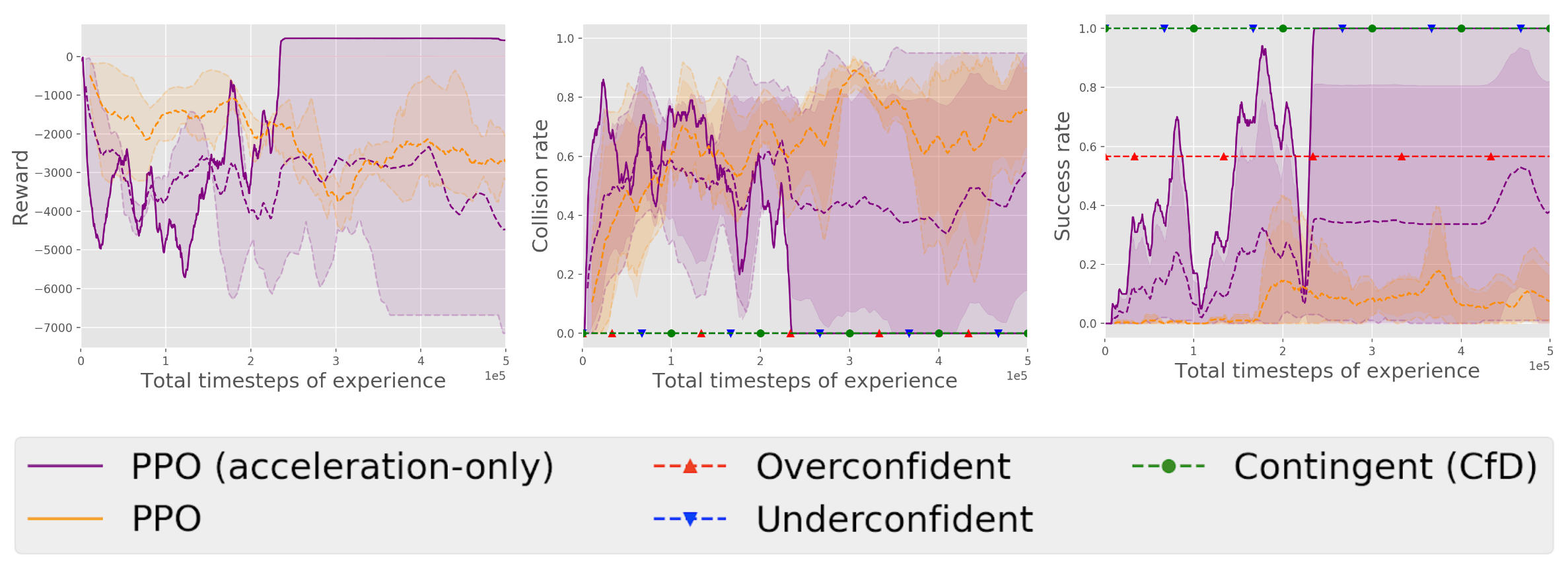}
    \vspace{-.1em}
    \caption{\small
    \textbf{Left:} Reward vs steps of experience on the left turn scenario (mean/std/min/max over three runs, best policy plotted in solid). 
    \textbf{Middle:} Collision rate vs steps. PPO is the only baseline that requires online training (all other methods are horizontal lines). PPO (acceleration only, similar to \cite{tang2019mfp}) achieves a 0\% collision rate. 
    \textbf{Right:} Success rate vs steps. PPO (acceleration only) achieves a 100\% success rate. PPO (throttle, steering and braking) is unable to exceed a 60\%.
    }
    \label{fig:modelfreecombined}
    \vspace{-1.75em}
\end{figure}

As seen in Fig. \ref{fig:modelfreecombined}, online model-free methods are capable of learning an optimal policy. 
However, the RL agent incurs a significant number of collisions before reaching a reasonable success rate
(in contrast, since \ours{} learns entirely offline, it need not experience collisions before learning how to complete the task).
Additionally, learning a successful policy with PPO requires careful reward shaping on top of the cost model used in \ours{}. Further details about the RL experimental setup, as well as example qualitative results of contingent and noncontingent plans, are given in the Appendix.

\vspace{-.4em}
\section{Discussion}
\vspace{-.35em}
We present a deep contingent planning method for control in multi-agent environments, and demonstrate its efficacy in common scenarios encountered in the urban driving setting in comparison to several alternative methods for data-driven planning, including a single-agent planner and a deep MPC-based trajectory shooting method. Our method is quite tractable to apply -- it is learned entirely from demonstrations of plausible multi-agent behavior and efficiently represents contingent plans in terms of components of the learned model.
 
\mypara{Acknowledgements}
This research was supported by the Office of Naval Research, ARL DCIST CRA W911NF-17-2-0181, JP Morgan, the Berkeley DeepDrive industry consortium, and the supporters of the Berkeley RISELab.

{ \footnotesize
\bibliographystyle{IEEEtranN}
\bibliography{references}
}

\clearpage
\appendix
\section{Appendix}

\subsection{Implementing the MFP baseline}

We were required to make several key changes to the publicly released MFP codebase to create a usable MFP baseline for our contingency planning scenarios in CARLA:
\begin{enumerate}
    \item The MFP codebase only includes code for the NGSIM dataset, so we added support for training (and visualizing) models on CARLA datasets, in addition to adding support for deploying the model in the CARLA simulator. Since MFP is a model for prediction (not planning), the latter required implementing a planning procedure to turn predictions from the MFP model into control for the ego vehicle (see Appendix \ref{sec:mfpplanning}).
    \item We added rotation of model inputs based on the each vehicle's yaw as an additional preprocessing stage, such that the reference frame for prediction always starts with the ego vehicle at (0,0) and pointing in the +X direction (the existing repository only shifts the inputs to (0,0)). We observed that rotation of inputs was necessary for training on the left and right turn scenarios.
    \item We removed the attention stage of the model architecture which significantly improved model performance on our CARLA datasets (recall that the scenarios we consider in this paper only include two agents).
    \item We removed the pretraining stage during model training (where the RNN/decoder output sequence is generated in a non-interactive fashion) which also significantly improved model performance.
\end{enumerate}
A repository containing a version of the MFP codebase with the aforementioned changes can be found on the project website. 

An important difference between \ours ~and MFP is that with \ours, we can train a single model on multiple scenarios, whereas with MFP, we are limited to training a single model for each individual scenario. We were unable to train MFP models on multiple scenarios that generated reasonable predictions (usable for planning): we found that the mixed-scenario models would often generate trajectories from the wrong scenario (e.g., a left turn on the right turn scenario). Additionally, we found it necessary to tune hyperparameters per-scenario, e.g., preprocessing with rotation is critical for the left and right turn scenarios, but degrades performance significantly on the overtake scenario.

\subsection{Planning with MFP}\label{sec:mfpplanning}

Similar to our behavior model described in \cref{sec:model_design}, MFP's forward pass (used during inference) generates predicted trajectories for $\bX_{\leq T}$.
Moreover, MFP generates $K$ different trajectories for each vehicle, where each trajectory $\bx_{\leq T}^{a,k}$ pertains to a different ``mode'' $k$ ($k \in 1, \dots, K$). 
According to the MFP authors, modes are meant to represent semantically different types of driving behavior such as intentions (e.g., left/right/straight) or behaviors (e.g., aggressive/conservative). 

MFP's multimodal prediction is implemented by using $K$ RNN decoders (one for each mode, with the total numbers of modes fixed during training), which model the future trajectories for all agents, conditioned on the past trajectories of all agents, the context view (the trajectories and visual context both processed by an encoder), and the mode: 
$p(\bY^a|\bX,\mathcal{I},k)$, 
paramaterized by a 2D Gaussian distribution over xy coordinates for each agent at every predicted timestep $t$, i.e., $\mathcal{N}_2(\mu,\rho)$ represented by the 5-tuple $(\mu^{a,t}_X$, $\mu^{a,t}_Y$, $\sigma^{a,t}_X$, $\sigma^{a,t}_Y$, $\rho^{a,t})$.
A given agent $a$'s history $\bX^a$ and an attention-based encoding of the history of other agents $Attn(\bX^{i \neq{} a})$ are concatenated in a vector $[\bX^a, Attn(\bX^{i \neq{} a}), \mathcal{I}]$ and passed as input to the encoding RNN, which produces a latent encoding and a Multinoulli distribution over modes per-agent, $p(K^a | \bX^a, Attn(\bX^{i \neq{} a}), \mathcal{I})$. The latent encoding is then input to each decoding RNN (one per mode) to generate $K$ predicted trajectories per agent.
All agents share the same set of (encoder and decoder) RNNs, so parameters scale linearly with modes, whereas the number of agents scales the batch/input dimension.

To summarize, the outputs of the trained MFP model represent the following distributions:

$p(\bY^a | \bX, \mathcal{I}, k)$, a distribution over future coordinates, per-agent, per-mode, generated by the decoding RNNs, and $p(K^a | \bX, \mathcal{I})$, a distribution over modes per-agent, generated by the encoding RNN. This means the model generates $K$ futures for each agent, i.e., $K \times A$ total trajectory distributions.

Note that the MFP model outputs do not include an explicit joint trajectory distribution $p(\bY | \bX, ...)$; the conditional joint future is factorized across agents, and across time (using the autoregressive decoder):
\begin{equation}\label{eq:mfpjoint}
    p(\bY^a | \bX, \mathcal{I}, k) = \prod_{t=1}^T \prod_{a=1}^A p(\by_t^a | \bY_{\leq t}, \bX, \mathcal{I}, k^a)
\end{equation}
We note that in Eq. \ref{eq:mfpjoint} (and its original formulation in \cite{tang2019mfp}), the joint is conditioned on the (ego) agent's own $k$, but not on the $k$s of other agents. However, in the reference implementation of MFP\footnote{\href{https://github.com/apple/ml-multiple-futures-prediction/commit/402f82d}{https://github.com/apple/ml-multiple-futures-prediction/commit/402f82d}} (and our own code), the joint is conditioned on all agents having the same $k$ value.

Planning using the same cost model as \ours ~requires explicit joint trajectories.
Since the MFP model outputs a distribution of $K$ trajectories for each of the $A$ agents, to generate candidate joint trajectories $\hat{\bY}$ in a two-agent setting, we can simply consider the set of all $K \times K$ possible combinations of individual ego and actor trajectories:
\begin{equation}
\begin{aligned}
    \hat{\bY}
    & = [(\mu_{\bY^{ego}}, \mu_{\bY^{actor}}) ~\text{for}~ k^{ego}, k^{actor} \in K]
    \\& = [(\mu^{ego}_X,\mu^{ego}_Y,\mu^{actor}_X,\mu^{actor}_Y) ~\text{for}~ k^{ego}, k^{actor} \in K]
\end{aligned}
\end{equation}
In our MFP experiments we used 5 modes, so we consider 25 total candidate joint trajectories. To select the best ego trajectory to use for planning, we can select the ego portion of the joint with the lowest cost (highest reward), weighted by the probability of each trajectory:
\begin{equation}
\begin{aligned}
    {\bY^{ego}}^\ast = 
    \argmin_{(\mu^{ego}_X,\mu^{ego}_Y)}~
    & (Cost(\mu^{ego}_X,\mu^{ego}_Y,\mu^{actor}_X,\mu^{actor}_Y)
    \\& \times p(K^{ego}=k^{ego} | \bX, \mathcal{I})
    \\& \times p(K^{actor}=k^{actor} | \bX, \mathcal{I}))
\end{aligned}
\end{equation}
Instead of weighting candidate joint trajectories by individual trajectory probabilities, alternatively we can filter out unrealistic joint trajectories by ignoring those with individual trajectory probabilities below a certain threshold, e.g.:
\begin{equation*}
    P(K^{ego}=k^{ego} | \bX, \mathcal{I}) < 0.01
\end{equation*}

Assuming a perfect prediction model, both of these methods will naturally create an \emph{overconfident} planner since we consider every possible combination of ego and actor trajectories. To create an \emph{underconfident} planner (assuming a perfect MFP model), we can choose the ego trajectory with the best worst-case joint trajectory, i.e., we assume an adversarial actor that chooses the actor trajectory leading to the worst performing joint, conditioned on the choice of ego trajectory:
\begin{equation}
\begin{aligned}
    {\bY^{ego}}^\ast = 
    & \argmin_{(\mu^{ego}_X,\mu^{ego}_Y)}~
     max_{(\mu^{actor}_X,\mu^{actor}_Y)}~
    \\& Cost(\mu^{ego}_X,\mu^{ego}_Y,\mu^{actor}_X,\mu^{actor}_Y)
\end{aligned}
\end{equation}

Finally, the form of the MFP model is amenable to contingent planning, in contrast to the action shooting originally described in \cite{tang2019mfp}. 

Once we have selected an optimal ego trajectory, we input the chosen ego trajectory to a PID controller to generate controls for the vehicle. In practice we found the outputs of MFP trained on the CARLA dataset to be imperfect, and observed that the underconfident planner is highly sensitive to noisy predictions and would often generate control that lead to out-of-distribution prediction. Therefore in our experiments we report results using the (weighted probability) overconfident planner.

\subsection{Implementing the model-free RL baseline} \label{app:model_free}
For our model-free reinforcement learning baseline, we use the implementation of Proximal Policy Optimization provided in the Stable Baselines repository~\citep{stable-baselines} (``PPO2''), and build on the OATomobile~\citep{filos2020can} repository to wrap each scenario in an OpenAI Gym environment.
The reward function used to train the RL agent is a linear combination of a collision penalty (to penalize risky behavior), constant timestep penalty (to encourage the agent to complete the turn as fast as possible), and ``lane-tracking'' reward (to encourage the agent follow a standard arc maneuver while turning). In practice we found the lane-tracking reward to be critical for learning realistic driving behavior.
We experiment with two different action spaces, one where the agent only controls the acceleration (steering is provided with an auto-pilot) which mirrors the RL experimental setup in \cite{tang2019mfp}, and another more realistic setting where the agent is also responsible for controlling steering and braking.
Each episode has a horizon of 200 steps but is terminated early if the agent completes the turn (i.e., reaches a target location).

We had to make several significant changes to the OATomobile codebase to enable RL training on our suite of contingency planning scenarios. Most importantly, the original codebase only allows control of a single agent in the CARLA simulator, i.e., the ego vehicle. However, our scenarios are multi-agent: the actor vehicle has to dynamically respond to the ego vehicle's state in order to implement the scenario behavior trees (e.g., Figure \ref{fig:unprotected_left_didactic}). To allow for training on a two-agent scenario, we modified the OATomobile API to take actor actions (in addition to ego actions) at each timestep, which are determined by the same behavior script described in Section \ref{sec:experiments}.

Although we were able to train an agent with PPO that manages to achieve a 0\% collision rate and 100\% success rate (using the restricted action space), we observe that training is highly unstable: safe policies were quick to diverge to degenerate solutions exhibiting highly overconfident behavior. We hypothesize that additional reward shaping and hyperparameter tuning may be able to stabilize training.

We were unable to learn a reasonable policy with RL using the larger (more realistic) action space, where the agent controls steering and braking in addition to acceleration. RL agents trained using the full action space generally exhibit erratic and unrealistic driving behavior. We hypothesize that additional reward shaping, post-processing of actions (e.g., smoothing in action space), and a pre-training stage (e.g., using expert data, or the restricted action space) may help in learning a more realistic policy.

\subsection{Visualizations of plans}

\begin{figure}[thb]
    \centering
    \includegraphics[width=.9\linewidth]{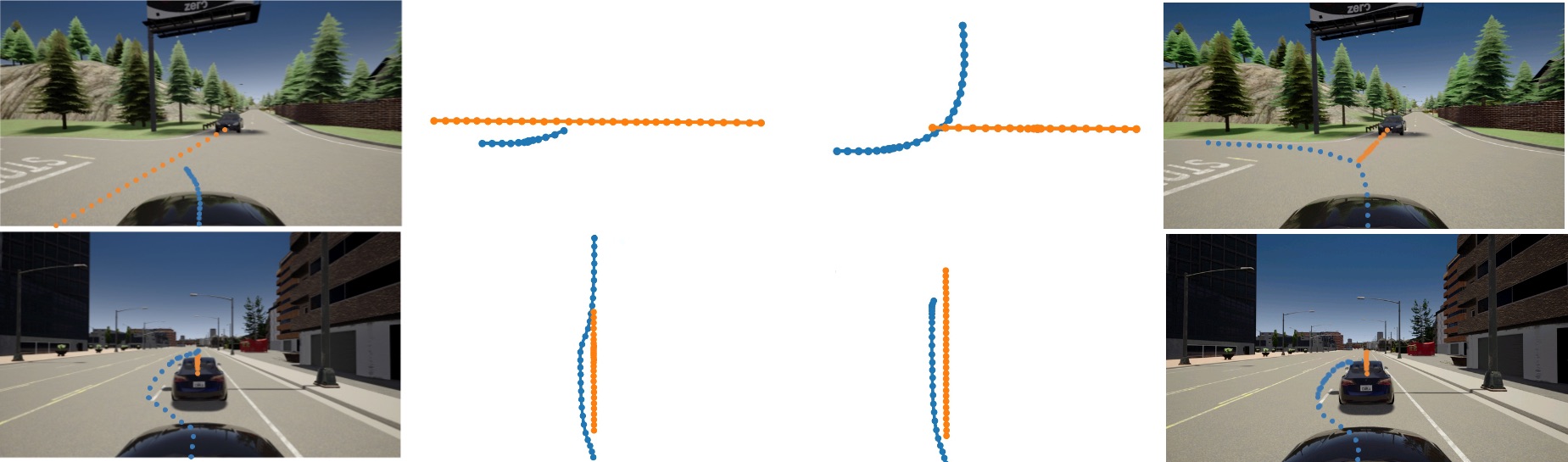}
    \caption{\small Examples of contingent plans in the left turn and overtake task. Each row corresponds to a contingent plan -- a single robot $\bz_{\leq T}^r$ -- and yields different joint trajectories when forward-simulated by the model with an accompanying sample of $\bz_{\leq T}^h$. Blue denotes the forecasted robot trajectory; orange denotes the forecasted human trajectory.}
    \label{fig:contingent_plan_examples}
    \vspace{-10pt}
\end{figure}

\begin{figure}[htb]
    \centering
    \begin{subfigure}[t]{.24\linewidth}
    \includegraphics[width=\linewidth]{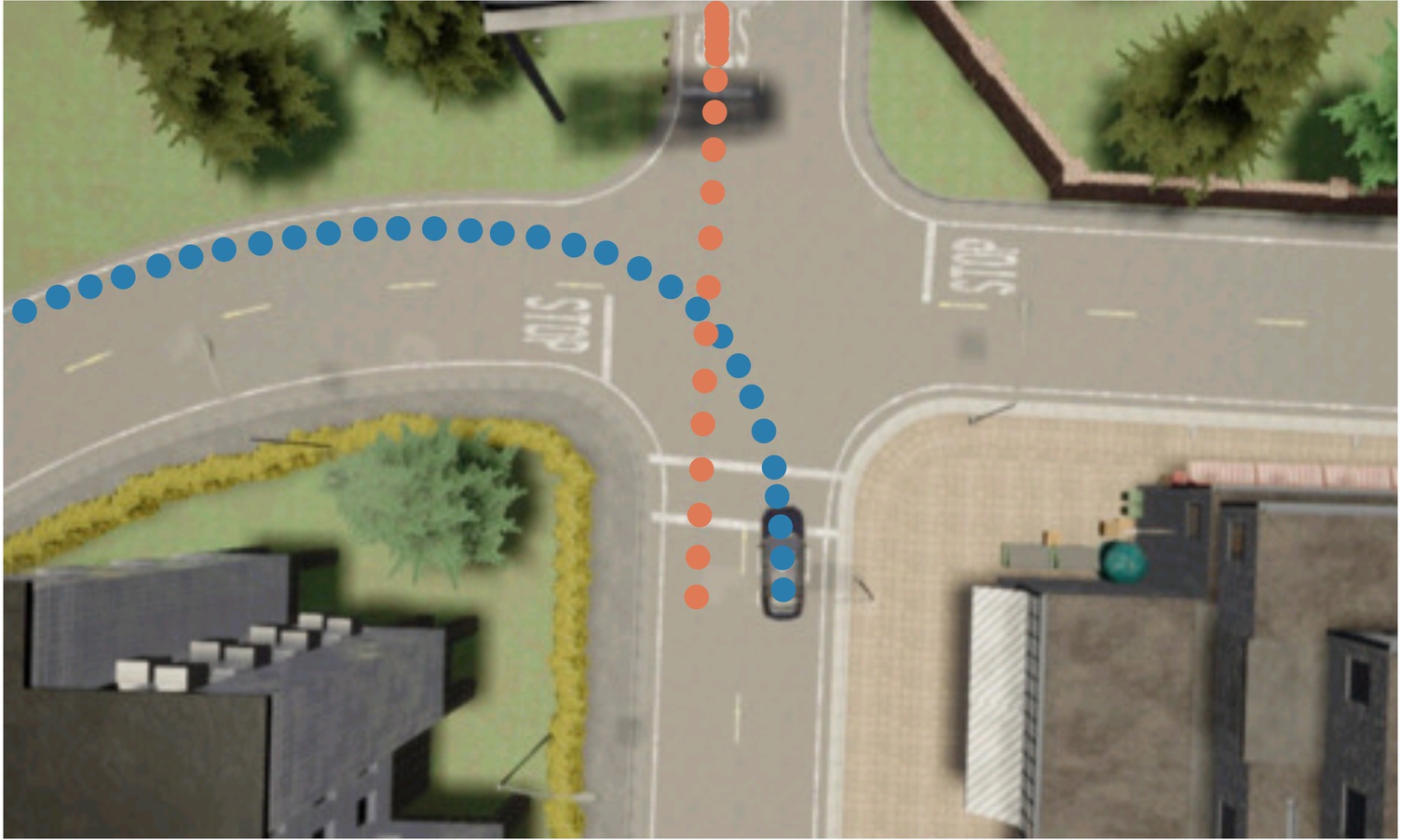}
    \end{subfigure}
    \begin{subfigure}[t]{.24\linewidth}
    \includegraphics[width=\linewidth]{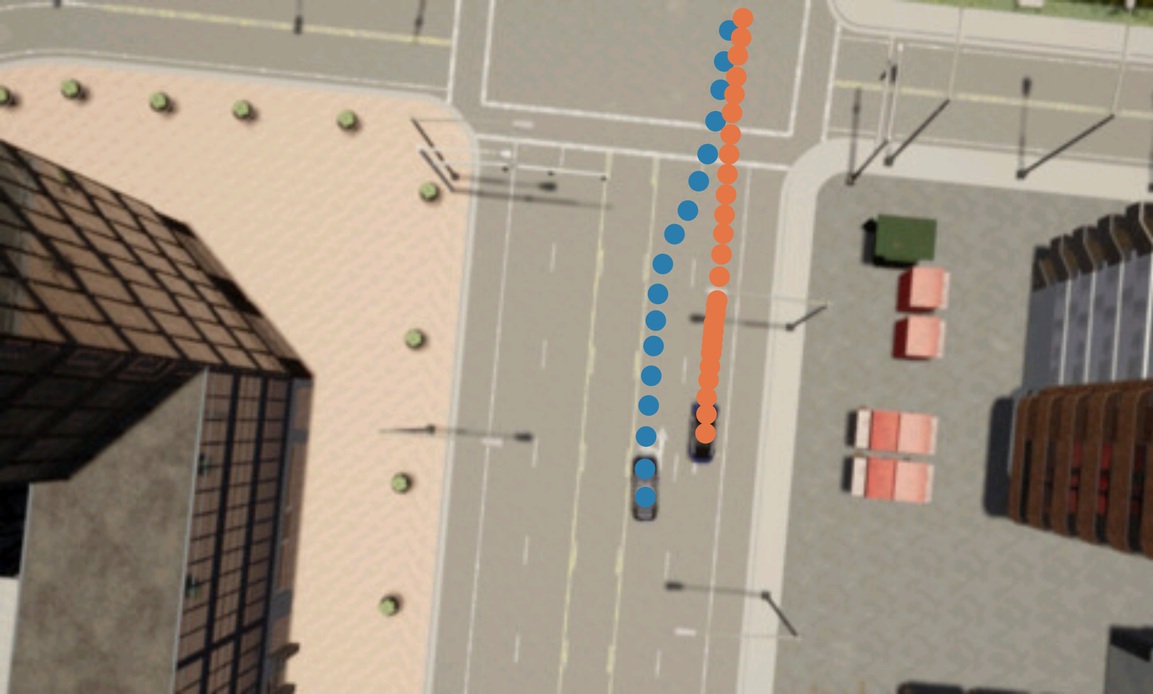}
    \end{subfigure}
    \begin{subfigure}[t]{.24\linewidth}
    \includegraphics[width=\linewidth]{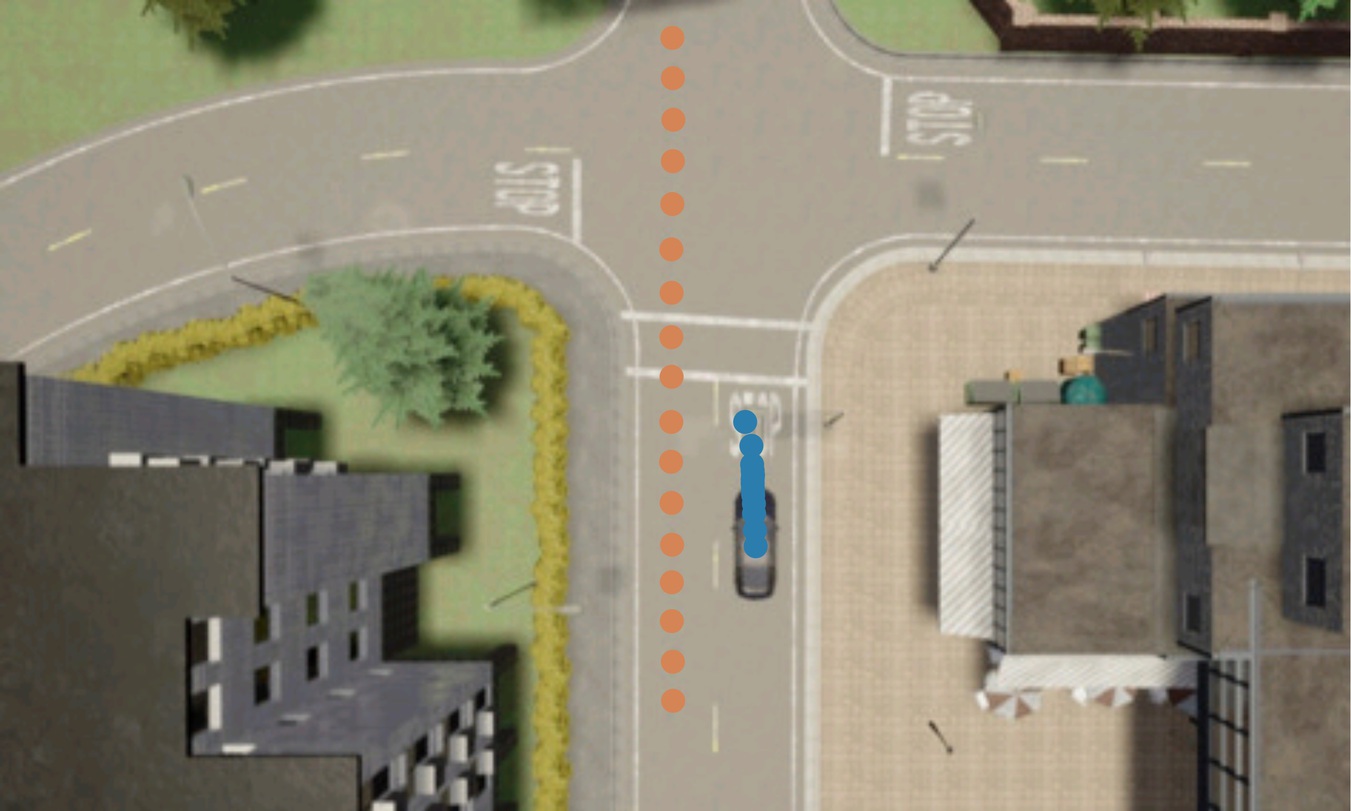}
    \end{subfigure}
    \begin{subfigure}[t]{.24\linewidth}
    \includegraphics[width=\linewidth]{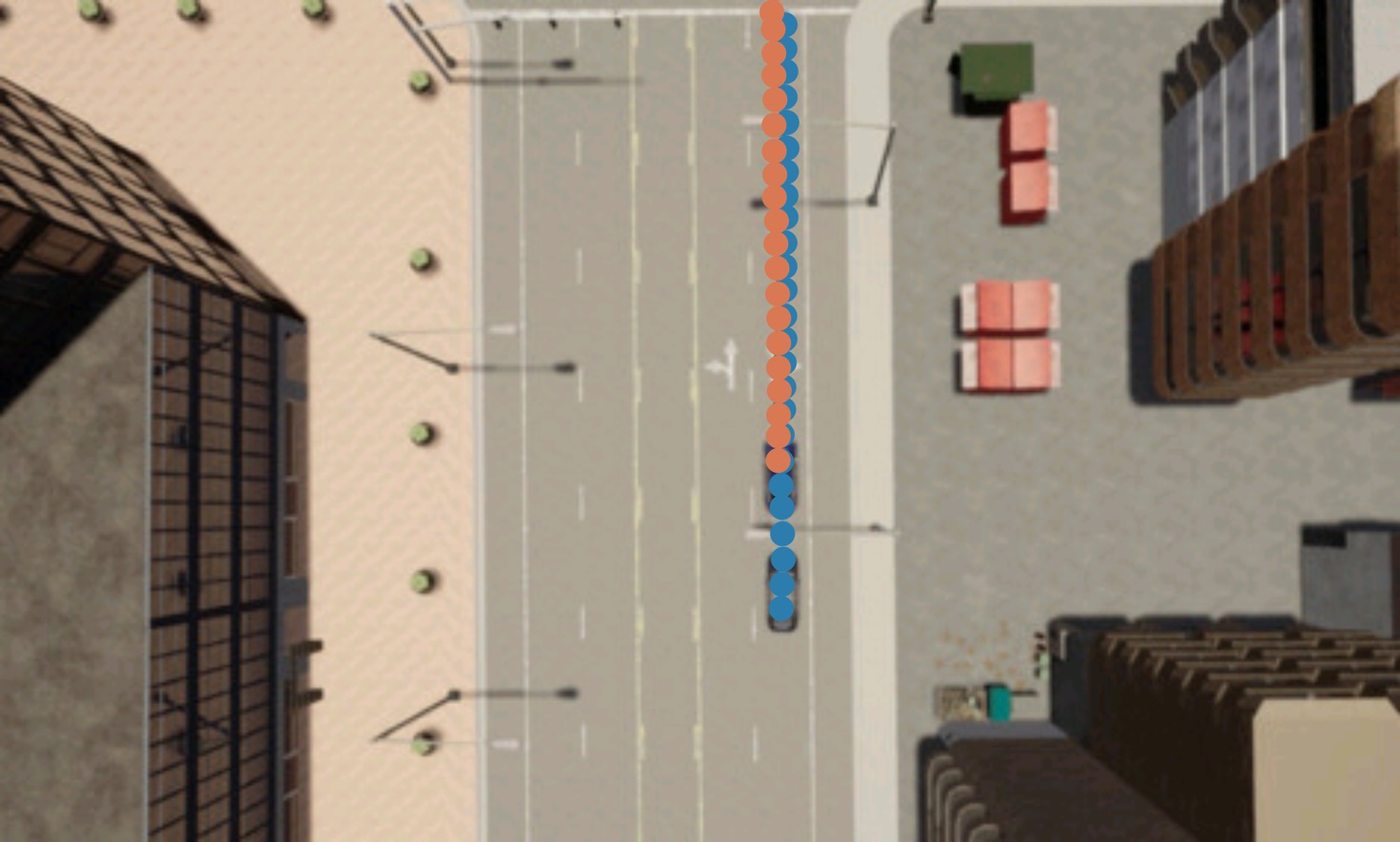}
    \end{subfigure} %
        \caption{\small Examples of overconfident (left half) and underconfident (right half) planning. In the left-turn and overtaking scenarios, the overconfident planner believes the human (orange) will always yield to the robot (blue). The underconfident planner believes that the robot cannot influence the human and plans a suboptimal determinstic path.}\label{fig:oc_uc}
    \vspace{-10pt}
\end{figure}

\subsection{Representing contingencies}
\begin{figure*}[htb]
 \centering
 \includegraphics[width=.9\textwidth]{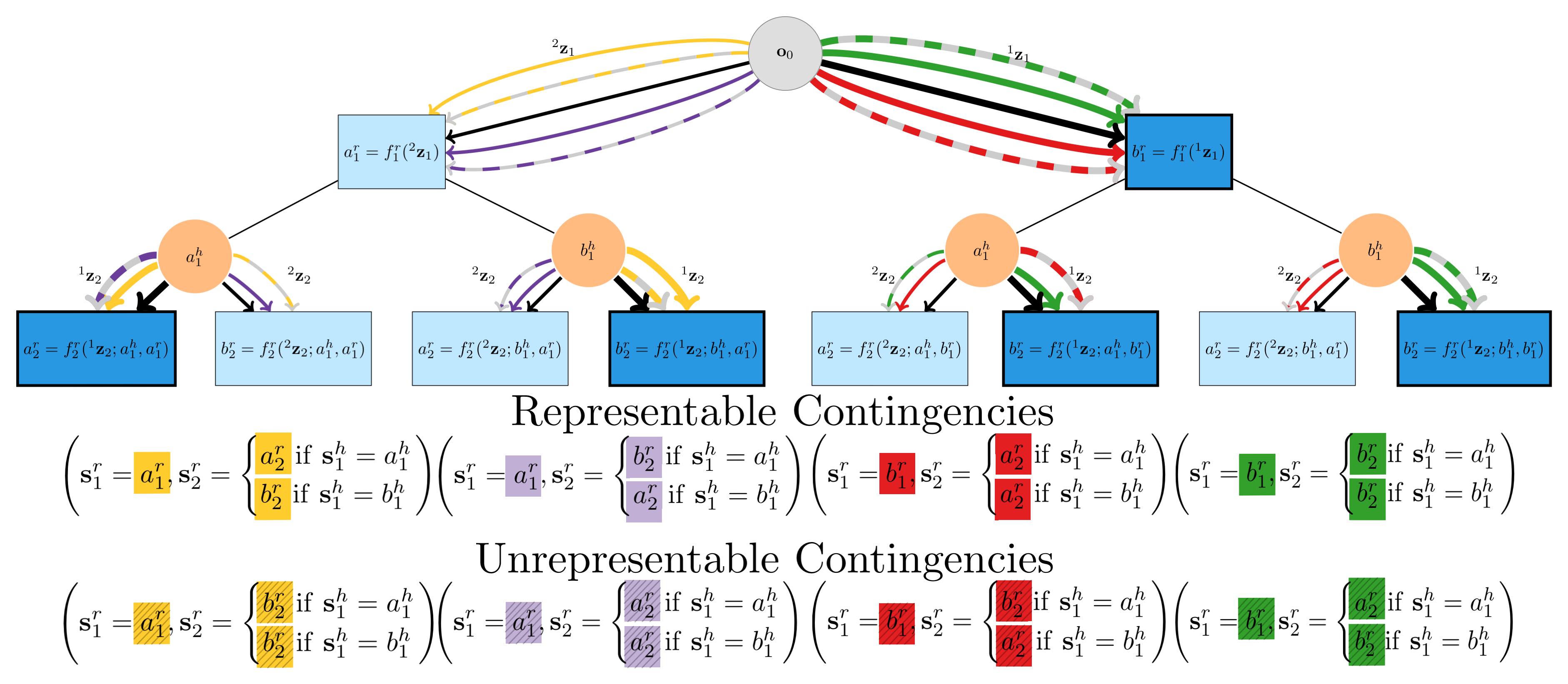}
\caption{A discrete autoregressive flow can represent contingency plans when provided with a subset of all base distribution values; the provided values represent the decisions that can be controlled/planned in a partially-controlled system. The represented contingencies include the most-likely responses at every time-step. In this case, the discrete autoregressive flow model is $q(\bX=\bx)\!=\!\prod_{t=1}^{T=2}\prod_{i=1}^{A=2}\bar{q}_t^i(\bZ_t^i=f_t^{-1,i}(\bx_t^i, \bx_{<t}))$. Agent $i$'s action at time $t$ is a random variable $\bX_t^i$ that assumes a value  $\bx_t^i \in \{a_t^i, b_t^i\}$, with $i=r$ denoting the robot (controlled agent), and $i=h$ denoting the human (uncontrolled agent). Blue nodes represent robot actions, orange nodes represent human actions, and dark blue nodes represent the likelier of two robot actions. Thick arrows denote the likelier of two actions under the model. Colored edges denote specific contingency plans; solid colored edges denote the representable contingencies, whereas gray-dashed colored edges denote the unrepresentable contingencies. By choosing the base-space reactions ahead of time ($\mathbf{z}_t^r$), this parameterization represents reactions ordered by their probabilities ${}^1\mathbf{z}_t^r=\argmax \bar{q}_t^r$, ${}^2\mathbf{z}_t^r=\argmax \bar{q}_t^r~\mathrm{s.t.}~{{}^2\bz_t^r \neq {}^1\mathbf{z}_t^r}\implies \bar{q}_t^r({}^1\mathbf{z}_t^r) > \bar{q}_t^r({}^2\mathbf{z}_t^r)$. Thus it cannot represent contingencies with model probability ordering that depend upon future data, but this is not a significant limitation. Note again that the representable contingencies include the
n-most-likely responses at every timestep.} \label{fig:binary_contingency_demo}
 \end{figure*}

 The model in Eq.~\ref{eq:esp_model} can be equivalently represented as an \emph{autoregressive flow} $q_f$, composed of learned bijections $f=[f_0^0, \dots, f_0^A, \dots, f_T^0, \dots, f_T^A]$ and base distributions $\bar{q}(\bZ)=\bar{q}_{\leq T}^{1:A}(\bZ_{\leq T}^{1:A})$. Consider the discrete setting. Discrete flows learn both the probability mass function $\bar{q}$ and $f$, and satisfy $q_f(\bX=\bx)\!=\!\bar{q}(\bZ\!\!=\!\!f^{-1}(\bx))$ \citep{tran2019discrete}; in our multi-agent autoregressive case, $\bar{q}(\bZ)\!=\!\prod_{t=1}^T\prod_{a=1}^A\bar{q}_t^a(\bZ_t^a)$. 
This implies that by using $\bz^r$ to parameterize contingency plans, the space of contingency plans that this can represent is the set of sequences of $n_t$-th ``best responses'' under the model.  
 To see this fact, let ${}^{n_t}\bz_t^r=\argmax_{\bz^t_r \in \mathbb Z^d \setminus \{{}^1\bz_t^r, \dots, {}^{n_t-1}\bz_t^r\}} \bar{q}_t^r$. When $n_t=1$, ${}^1\bz_t^r$ is the first best response at time $t$, as judged by $\bar{q}_t^r$. When $n_t>1$, ${}^{n_t}\bz_t^r$ is the $n_t$-th best response at time $t$. Thus, at every time step $t$, a choice of $\bz_t^r$ is identified with the $n_t$-th best response ${}^n\bz_t^r$. Thus, the space of contingent plans is $\{{}^{n_{1}}\bz_{1}^r, \dots, {}^{n_{T}}\bz_{T}^r : n_{t} \in \mathbb N\}$; the best contingency plan under the model is ${}^{1,\dots,1}\bz^r_{\leq T}$. Therefore, the space of contingent plans that a discrete autoregressive flow model can represent is relatively flexible, as the contingent plans that this parameterization does not represent are just those that have values of $n_t$ that are contingent on the possible futures $\bX_{1:t-1}$ (e.g. policies that perform the first best response to partial future $X$ and the second best response to partial future $Y$).  Although our model is not discrete, these observations lead us to speculate
 that there generally will exist $\bz_{\leq T}^r$ for each scene that represent high-quality contingent plans; evidence from our experiments support this speculation. An illustration of the contingencies that a binary autoregressive flow can represent with $\bz_{\leq T}^r$ is shown in Fig.~\ref{fig:binary_contingency_demo}.

\subsection{Planning criterion derivation}
We derive the planning criterion of Eq.~\ref{eq:final_planning_criterion} as a MAP estimation of a scene-conditioned posterior of policy parameters, $p(\bz_{\leq T}^r|\mathcal G, {\bo})$.
\renewcommand{\CancelColor}{\color{gray}}
{
\begin{align}
    &\argmax_{\bz_{\leq T}^r} p(\bz_{\leq T}^r|\mathcal G, {\bo}) \nonumber \\
    &=\argmax_{\bz_{\leq T}^r} \log \, p(\mathcal G,\bz_{\leq T}^r|{\bo}) \,-\,
      {\log p(\mathcal G|{\bo})} 
      \label{eq:planning_criterion_first_step} \\
      &=\argmax_{\bz_{\leq T}^r} \!\! \log \!\! \int \!\! p(\mathcal G|\bz_{\leq T}^r,\!\bz^h\!\!\!, {\bo}\!)p(\bz_{\leq T}^r,\!\bz^h|{\bo}\!)p(\bz^h\!|{\bo}\!)\mathrm{d}\bz^h \label{eq:planning_criterion_middle_step}\\
      &=\argmax_{\bz_{\leq T}^r} \log \mathbb E_{p({\bz}^h|{\bo})} p(\mathcal G|\bz_{\leq T}^r,\bz^h, {\bo})p(\bz_{\leq T}^r,\bz^h|{\bo})  \label{eq:planning_criterion}
\end{align}
}%
where between \cref{eq:planning_criterion_first_step} and \cref{eq:planning_criterion_middle_step} we drop the term $\log p (\mathcal{G} | {\bo})$ since it is independent of $\bz_{\leq T}^r$.
Now, we incorporate the learned prior of expert behavior $q_\theta(\bX\!=\!\bx|{\bo})$
by defining the prior over policy parameters and human stochasticity $p(\bz_{\leq T}^r, \bz^h|{\bo})\!\doteq\! q_\theta(\bX\!=\!f(\bz^r_{\leq T}, \bz^h)|{\bo})$. We incorporate the goals $\mathcal G$ in terms of two distributions: $p(\mathcal G|\bz_{\leq T}^r,\bz^h, {\bo})\doteq p(\mathbf{g}|\bz_{\leq T}^r,\bz^h,{\bo})\delta_{\mathbb G}(\bx)$, with $p(\mathbf{g}|\bz_{\leq T}^r,\bz^h, {\bo})\doteq \mathcal N(f(\pi^r_{\bz_{\leq T}}, \bz^h)_T^r; \mathbf{g}, I)$ -- this favors $\pi^r_{\bz_{\leq T}^r}$ that tend to end near $\mathbf{g}$. Finally, we use $p(\bz^h|{\bo})\doteq \mathcal N(0, I)$, which is the model's prior over human stochasticity. In practice, we perform stochastic gradient ascent on a numerically version of the criterion in Eq.~\ref{eq:planning_criterion} that swaps the ordering of $\log$ and $\mathbb E$ (by Jensen's inequality) to arrive at a lower-bound, and approximates a violated constraint ($\log \delta_{\mathbb G}(\bx)\!=\!-\infty$) with a large negative value. This final planning criterion is used to implement \oursfull~(\ours), and is reproduced in Eq.~\ref{eq:app_final_planning_criterion}.

 \begin{align}
    &\mathcal{L}_\text{\ours}(\pi^r_{\bz_{\leq T}})
    =  \label{eq:app_final_planning_criterion}\\
    &\mathop{\E}_{\bz^h \sim \mathcal{N}(0, I)} \Big[
    \log \underbrace{q_\theta(\bar{\bx}_{\leq T}|{\bo})}_{(1)~\text{prior}}\!+\!\log  \underbrace{\mathcal N\big(\bar{\bx}^r_T; \mathbf{g}, I
    \big)}_{(2)~\text{destination}}\!+\!\log \underbrace{\delta_{\mathbb G}(\bar{\bx}_{\leq T})}_{(3)~\text{constraint}} 
    \Big].\notag
 \end{align}

\subsection{Theoretical Justification for Contingency Planning}
Here we demonstrate why value increases if our planning considers contingencies, rather than only planning a single state sequence. We additionally show the benefit of active planning, where the robot is known to affect other agents.
Consider planning just two time steps into the future as visualized in \cref{fig:models}. Both the ``No leader'' and ``Robot leader'' methods plan full state trajectories open-loop, with associated value functions $Q^{\text{NL}}$ and $Q^{\text{RL}}$. By contrast, the contingent planners (``Human leader'' and ``Co-leaders'', with values $Q^{\text{HL}}$ and $Q^{\text{CL}}$) understand that future decisions like $\bx^r_2$ depend on future information, like the value of $\bx^h_1$, which is not yet known. Below we show the value functions corresponding to each planner. We use the hat symbol \smash{$\hat{Q}$} to reflect the estimated value function that each planner optimizes, an estimate which is based on different modelling assumptions about how multi-agent interaction occurs:

\newcommand{\reward}[1]{\text{reward}(\bx^r_#1,\bx^h_#1)}
\newcommand{\objectivereturn}[0]{\text{return}(\bx^{r,h}_{1,2})}
\begin{alignat}{2}
\hat{Q}^{\text{NL}}({\bo},\bx^r_1) &\doteq \max_{\bx^r_2} \, \blue{\mathbb{E}_{\bx^h_1|{\bo}}} \, \orange{\mathbb{E}_{\bx^h_2|\bx^h_1,{\bo}}} && \Big[\objectivereturn\Big], 
\label{eq:vn} \\
\hat{Q}^{\text{RL}}({\bo},\bx^r_1) &\doteq \max_{\bx^r_2} \, \blue{\mathbb{E}_{\bx^h_1|{\bo}}} \, \green{\mathbb{E}_{\bx^h_2|\bx^r_1,\bx^h_1,{\bo}}} && \Big[\objectivereturn\Big], 
\label{eq:vr} \\
\hat{Q}^{\text{HL}}({\bo},\bx^r_1) &\doteq \blue{\mathbb{E}_{\bx^h_1|{\bo}}} \, \max_{\bx^r_2} \, \orange{\mathbb{E}_{\bx^h_2|\bx^h_1,{\bo}}} && \Big[\objectivereturn\Big], 
\label{eq:vh} \\
\hat{Q}^{\text{CL}}({\bo},\bx^r_1) &\doteq \blue{\mathbb{E}_{\bx^h_1|{\bo}}} \, \max_{\bx^r_2} \, \green{\mathbb{E}_{\bx^h_2|\bx^r_1,\bx^h_1,{\bo}}} && \Big[\objectivereturn\Big].
\label{eq:vc}
\end{alignat}

\paragraph{Case for \textit{contingency} planning}
For any $({\bo},\bx^r_1)$, we have $\hat{Q}^{\text{NL}} \leq \hat{Q}^{\text{HL}}$ and $\hat{Q}^{\text{RL}} \leq \hat{Q}^{\text{CL}}$ due to Jensen's inequality, noting the maximum operator is convex. 
More intuitively, by moving a max operator after the blue expectation operator, the value function can realize the ``best of all worlds'', by planning optimal contingencies for each future outcomes. By contrast, a max operator before the expectation computes evaluates the single best ``compromise'' decision $\bx^r_2$ that is common to all possible values of the preceding human state $\bx^h_1$.

\paragraph{Case for \textit{active} contingency planning}
Note $\hat{Q}$ denotes the value function that each planner aims to optimize, but is not necessarily the ``true value'' of such a planner (denoted $Q$) if $\hat{Q}$ was based on erroneous assumptions.
In \cref{eq:vn,eq:vr,eq:vh,eq:vc} the orange expectation operator is based on a incomplete behavior model that predicts $\bx^h_2$ given $(\bx^h_1,{\bo})$, even though $(\bx^h_1,{\bo})$ is not the Markov state, since it omits the robot's influence on the human. The full Markov state to predict the next human state is featured in the green expectation operator, which assumes a behavior model to predict $\bx^h_2$ given all the information: $(\bx^r_1,\bx^h_1,{\bo})$. We thus consider the green operator to reflect a behavior model that is more complete, and faithful to how reality will progress, compared to the orange operator. Thus, we have $\hat{Q}^{\text{RL}}=Q^{\text{RL}}$ and $\hat{Q}^{\text{CL}}=Q^{\text{CL}}$, but the same cannot generally be said for $\hat{Q}^{\text{NL}}$ and $\hat{Q}^{\text{HL}}$.

With respect to the random variable of the orange expectation, we can expand it:
\begin{align*}
p(\bx^h_2|\bx^h_1,{\bo}) &= \int\! p(\bx^h_2,\bx^{r'}_1\!|\bx^h_1,{\bo}) \text{d}\bx^{r'}_1 \\
&= \int\! p(\bx^h_2|\bx^{r'}_1\!,\bx^h_1,{\bo}) p(\bx^{r'}_1\!|{\bo}) \text{d}\bx^{r'}_1 \\
&=\, \mathbb{E}_{\bx^{r'}_1\!|{\bo}}\!\left[p(\bx^h_2|\bx^{r'}_1\!,\bx^h_1,{\bo})\right].
\end{align*}
Thus we can represent $\orange{\mathbb{E}_{\bx^h_2|\bx^h_1,{\bo}}} = \red{\mathbb{E}_{\bx^{r'}_1\!|{\bo}}} \yellow{\mathbb{E}_{\bx^h_2|\bx^{r'}_1\!,\bx^h_1,{\bo}}}$
where $\bx^{r'}_1$ represents the robot state that was implicitly integrated-out at training time, instead of being explicitly conditioned on. This implicit variable $\bx^{r'}_1$ is distinct from the robot-state at \textit{test-time} $\bx^r_1$ which the planner optimizes. 
This way of re-framing \cref{eq:vh} frames the human-leader model as omitting the known (planned) value of $\bx^r_1$ at test-time when predicting $\bx^h_2$ during planning, and instead reverting to the prior distribution of robot states seen at training times: $p(\bx^{r'}_1\!|{\bo})$. Omitting known information reduces the value of such a planner. The ``true'' value of the human-leader planner can be computed using $Q^{\text{CL}}$ given $Q^{\text{CL}}$'s definition assumes the correct process (green operator) of how $\bx^h_2$ is generated:
\begin{align}
V^{\text{HL}}({\bo})
&= Q^{\text{CL}}\big({\bo},\,\argmax_{\bx^r_1} \hat{Q}^{\text{HL}}({\bo},\bx^r_1)\big) \\
&\leq Q^{\text{CL}}\big({\bo},\,\argmax_{\bx^r_1} Q^{\text{CL}}({\bo},\bx^r_1)\big)
= V^{\text{CL}}({\bo}),  \nonumber
\end{align}
where we define a corresponding $V$-function for each $Q$-function in \cref{eq:vn,eq:vr,eq:vh,eq:vc}:
\begin{align}
V({\bo}) \;&\doteq\; \max_{\bx^r_1} \, Q({\bo},\bx^r_1).
\end{align}
By analogous reasoning, $V^{\text{NL}}({\bo}) \leq V^{\text{RL}}({\bo})$, meaning \textit{active} planning is beneficial among non-contingent planners too: whenever the robot considers how its actions affect the world, and those effects have relevant repercussions in the form of objective rewards, the planner can simulate how robot actions can such repercussions to achieve a higher return.

\paragraph{Summary}
The benefit of (1) \textit{contingency} planning, and (2) \textit{active} planning discussed above, results in a partial ordering of the \textit{true} values of each planning category seen in \cref{fig:models}:
\begin{align}
V^{\text{NL}} \;\;\leq\;\; \{V^{\text{RL}},V^{\text{HL}}\} \;\;\leq\;\; V^{\text{CL}},
\end{align}
noting the value of active contingency planning is greater-or-equal to all other planning methods.

\paragraph{T-step $Q$-values}

In a more general horizon of $T$ time-steps, the various $Q$-values are below. For brevity, we do not show the conditioning on $\bo$, and treat it as implicit.

\newcommand{\fullreturn}[0]{\text{return}(\bx^{r,h}_{1:T})}

\small{
\begin{alignat}{2}
\hat{Q}^{\text{NL}}(\bx^r_1) &\doteq \max_{\bx^r_2} \max_{\bx^r_3} \cdots \max_{\bx^r_T} \, \blue{\mathbb{E}_{\bx^h_1}} \orange{\mathbb{E}_{\bx^h_2|\bx^h_1}} \cdots \mathbb{E}_{\bx^h_T|\bx^h_{1:T-1}} && \Big[\fullreturn\Big], \nonumber \\
\hat{Q}^{\text{RL}}(\bx^r_1) &\doteq \max_{\bx^r_2} \max_{\bx^r_3} \cdots \max_{\bx^r_T} \blue{\mathbb{E}_{\bx^h_1}} \green{\mathbb{E}_{\bx^h_2|\bx^{r,h}_1}} \cdots \mathbb{E}_{\bx^h_T|\bx^{r,h}_{1:T-1}} && \Big[\fullreturn\Big], \nonumber \\
\hat{Q}^{\text{HL}}(\bx^r_1) &\doteq \blue{\mathbb{E}_{\bx^h_1}} \max_{\bx^r_2} \orange{\mathbb{E}_{\bx^h_2|\bx^h_1}} \max_{\bx^r_3} \cdots \max_{\bx^r_T} \mathbb{E}_{\bx^h_T|\bx^h_{1:T-1}} && \Big[\fullreturn\Big], \nonumber \\
\hat{Q}^{\text{CL}}(\bx^r_1) &\doteq \blue{\mathbb{E}_{\bx^h_1}} \max_{\bx^r_2} \green{\mathbb{E}_{\bx^h_2|\bx^{r,h}_1}} \max_{\bx^r_3} \cdots \max_{\bx^r_T} \mathbb{E}_{\bx^h_T|\bx^{r,h}_{1:T-1}} && \Big[\fullreturn\Big]. \nonumber
\end{alignat}
}

\end{document}